\documentclass[a4paper,fleqn]{cas-dc}

\usepackage{natbib}
\usepackage{amsmath}
\usepackage{algorithm}
\usepackage{mathtools}
\usepackage{array}
\usepackage{multirow}
\usepackage{multicol}
\usepackage[caption=false,font=normalsize,labelfont=sf,textfont=sf]{subfig}
\usepackage{url}
\usepackage{amsthm}

\usepackage[noabbrev]{cleveref}
\usepackage{amsfonts}

\usepackage{subfig}
\usepackage{mwe}

\newtheorem{definition}{Definition}
\usepackage{algpseudocode} 

\usepackage{enumitem}
\setlist[enumerate]{leftmargin=*}
\setlist[itemize]{leftmargin=*}

\begin{document}
\let\WriteBookmarks\relax
\def\floatpagepagefraction{1}
\def\textpagefraction{.001}

\title [mode = title]{SE-shapelets: Semi-supervised Clustering of Time Series Using Representative Shapelets}                      

\tnotetext[1]{Corresponding author}

%
\author[1]{Borui Cai}

\cormark[1]

\author[1]{Guangyan Huang}

\author[2]{Shuiqiao Yang}

\author[1]{Yong Xiang}

\author[3]{Chi-Hung Chi}
\affiliation[1]{organization={School of Information Technology, Deakin University},
    city={Burwood},
    postcode={3125}, 
    state={VIC},
    country={Australia}}

\affiliation[2]{organization={School of Computer Science and Engineering, University of New South Wales},
    city={Sydney},
    postcode={2032}, 
    state={NSW},
    country={Australia}}

\affiliation[3]{organization={Data61 in CSIRO},
    city={Hobart},
    postcode={7005}, 
    city={TAS},
    country={Australia}}

\nonumnote{\textit{Email addresses}: b.cai@deakin.edu.au (B.Cai); \newline
guangyan.huang@deakin.edu.au (G.Huang);
shuiqiao.yang@unsw.edu.au (S.Yang); 
yong.xiang@deakin.edu.au (Y.Xiang); \newline
chihung.chi@csiro.au (CH.Chi)
  }

\pagebreak

\begin{abstract}
Shapelets that discriminate time series using local features (subsequences) are promising for time series clustering. Existing time series clustering methods 
may fail to capture representative shapelets because they discover shapelets from a large pool of uninformative subsequences, and thus result in low clustering accuracy. This paper proposes a Semi-supervised Clustering of Time Series Using Representative Shapelets (SE-Shapelets) method, which utilizes a small number of labeled and propagated pseudo-labeled time series to help discover representative shapelets, thereby improving the clustering accuracy. In SE-Shapelets, we propose two techniques to discover representative shapelets for the effective clustering of time series. 1) A \textit{salient subsequence chain} ($SSC$) that can extract salient subsequences (as candidate shapelets) of a labeled/pseudo-labeled time series, which helps remove massive uninformative subsequences from the pool. 2) A \textit{linear discriminant selection} ($LDS$) algorithm to identify shapelets that can capture representative local features of time series in different classes, for convenient clustering. Experiments on UCR time series datasets demonstrate that SE-shapelets discovers representative shapelets and achieves higher clustering accuracy than counterpart semi-supervised time series clustering methods.
\end{abstract}


\maketitle

\section{Introduction}
Time series is an important data type and can be collected from pervasive scenarios, ranging from wearable devices \citep{wear} and sensory systems \citep{sensor} to autonomous vehicles \citep{intro_radar,uav1}. Clustering is a fundamental tool and plays an essential role in multiple time series tasks, such as data preprocessing \citep{process}, image segmentation \citep{seg}, and pattern recognition \citep{sensory1}.
Shapelets \citep{ushape} are promising for effective time series clustering by capturing subsequence/local features to discriminate time series of different classes. For time series clustering, existing methods \citep{ushape,ussl} normally discover several shapelets that capture representative local features of time series, and then map time series to distance-to-shapelets representations for the clustering.

\begin{figure}[!t]
\centering
\includegraphics[width=3.3in,height=1.4in]{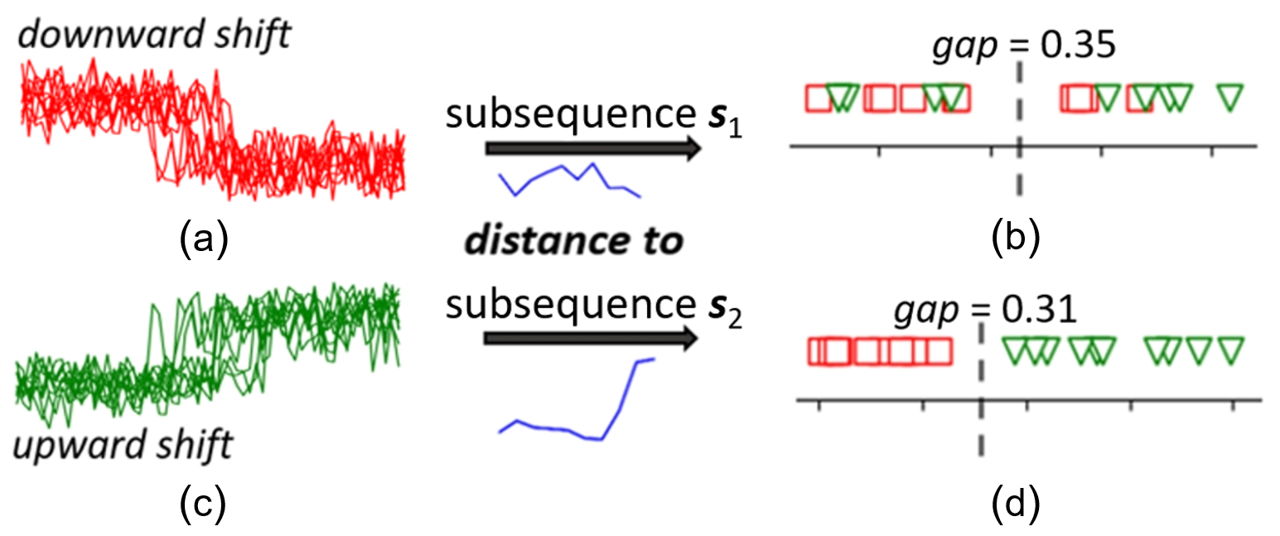}
\caption{(a) and (c) show time series of two shift patterns, and their distance distributions with two candidate shapelets ($\boldsymbol{s}_{1}$ and $\boldsymbol{s}_{2}$) are shown in (b) and (d), respectively. Although $\boldsymbol{s}_{2}$ captures the representative local pattern of $upward\ shift$ and can discriminate two types of time series, its $separation\ gap$ is worse than that of $\boldsymbol{s}_{1}$ (causes incorrect groups).}
\label{fig:intro}
\end{figure}

Discovering representative shapelets is essential for the effectiveness of time series clustering. Such shapelets can map time series of the same classes into distinct and compact groups, which are convenient to be clustered \citep{ushape}. To discover representative shapelets, existing methods normally develop specific quality measurements to select high-quality time series subsequences as shapelets, from a large pool of subsequences \citep{ushape,sd}. Existing shapelet quality measurements mainly adopt statistical analysis on the distance-to-shapelet distribution of time series with a candidate shapelet. However, since no prior information of the dataset is available, they cannot measure if such distribution satisfies the classes of time series. For example, the $separation\ gap$ \citep{ushape} prefers the subsequence that produces two far-away groups in the distribution, but does not know if these groups have high purity. That means, the discovered shapelets may not well discriminate time series of different classes and result in unsatisfactory clustering accuracy. We show an example in Fig. \ref{fig:intro} using SyntheticControl dataset \citep{synthetic}. Fig. \ref{fig:intro} (a) and (c) show time series of two different shift patterns, and they are respectively mapped to distance-to-shapelets representations with two candidate shapelets ($\boldsymbol{s}_{1}$ and $\boldsymbol{s}_{2}$), as shown in Fig. \ref{fig:intro} (b) and (d). 
From the distributions, $\boldsymbol{s}_{2}$ produces a $separation\ gap$ (0.31) worse than that of $\boldsymbol{s}_{2}$ (0.35), though $\boldsymbol{s}_{2}$ captures the representative local pattern of $upward\ shift$ and correct discriminates the two different types of time series.

Similar kinds of problems in other fields have been resolved by semi-supervised clustering approaches, which use slight supervision \citep{cssc} (e.g., a small number of labeled data objects or manual constraints) to improve the clustering performance. 
For example, SemiDTW determines the optimal window size for the DTW distance with the support of a small number of manual constraints, and thus improves the performance of time series clustering \citep{semidtw}.
Inspired by this, our vision is to perform effective semi-supervised time series clustering using representative shapelets, which are discovered with a small number of labeled time series and the propagated pseudo-labeled time series.
So, two problems need to be properly addressed. First, how to prepare a limited number of informative subsequences as shapelet candidates is unclear. Discovering shapelets from all possible subsequences of the labeled and propagated pseudo-labeled time series may result in over-fitting and is also computationally expensive, due to the large size of uninformative subsequences (e.g., stop-word subsequences \citep{stopword}). Although some existing methods attempt to prune similar/repetitive subsequences \citep{sd} or directly use random subsequences \citep{random}, it remains untouched what type of subsequence is suitable to be considered as shapelet candidates.
Second, how to utilize labels to discover shapelets for time series clustering is unexplored. Existing shapelet quality measurements for time series clustering (e.g., $separation\ gap$) cannot be directly adopted since it does not consider the purity of formed groups. While shapelet quality measurements that utilize dataset information (the labels of time series), e.g., \textit{nearest neighbour accuracy} \citep{sd} and \textit{information gain} \citep{gain0} are designed for time series classification and may not discover optimal shapelets for clustering.

In this paper, we propose a Semi-supervised Clustering of Time Series Using Representative Shapelets (SE-shapelets) method for accurate time series clustering. Different from unsupervised shapelet-based time series clustering methods, e.g., U-shapelets \citep{ushape}, SE-shapelets performs time series clustering by discovering representative shapelets with a small number of labeled time series and pseudo-labeled time series (propagated from nearest labeled time series). In SE-shapelets, we provide two techniques to address the two aforementioned problems. First, inspired by the definition of ``salience'' in neuroscience (contrasts between items and their neighborhood) \citep{salience}, we define a new \textit{salient subsequence chain} ($SSC$) to extract a small number of subsequences representing salient local features from a time series; only these salient subsequences are considered as shapelet candidates, to avoid massive uninformative subsequences. Second, we propose a \textit{linear discriminant selection} ($LDS$) algorithm to select shapelets that can capture representative local features (based on labels/pseudo-labels) for clustering. Specifically, $LDS$ tends to select shapelets that can map time series of the same classes into distinct and compact groups, for convenient clustering.

In summary, this paper makes the following contributions:
\begin{itemize}
\item[--] We propose a SE-shapelets method to improve the accuracy of time series clustering, by discovering representative shapelets with a small number of labeled and pseudo-labeled time series.
\item[--] We propose \textit{salient subsequence chain} ($SSC$) to extract salient subsequences from a time series, and develop an efficient $FindChain$ algorithm to discover $SSC$.
\item[--] We propose a \textit{linear discriminant selection} ($LDS$) algorithm, which can utilize labels/pseudo-labels to discover representative shapelets.
\item[--] We conduct comprehensive experiments to evaluate the proposed SE-shapelets. The results show that SE-shapelets achieves promising clustering accuracy that surpasses state-of-the-art counterpart methods.
\end{itemize}

The rest of this paper is organized as follows. The related works are reviewed in Section II. The preliminary knowledge is introduced in Section III. The proposed SE-shapelets is detailed in Section IV and evaluated in Section V. The paper is summarized in Section VI.

\section{Related Work}
Time series clustering has been a hot research field for decades due to its pervasive applications. In this section, we briefly review the existing time series clustering methods and semi-supervised time series clustering methods.

\subsection{Time Series Clustering}
\textbf{Conventional Methods.} Time series clustering aims at grouping similar time series into the same group, while separating distinctive time series into different groups \citep{survey}. A main challenge is that time series widely suffers from various distortions (e.g., phase shifting, time warping \citep{minidtw}), and many methods are proposed to address this problem. Dynamic time warping (DTW) is a distance measurement that can find the optimal alignment of time series, and KDBA \citep{kdba} extends Kmeans, by adopting a global averaging technique, to enable the use of DTW distance measurement for time series clustering. To avoid the high time complexity of DTW, Kshape \citep{kshape} develops an effective shape-based distance (SBD) for time series, while YADING \citep{yading} adopts L1-Norm to discover natural-shaped clusters by analyzing the density distribution. Other distance measurements, such as longest common subsequence \citep{longest}, edit distance \citep{edit}, and shape-based distance \citep{kshape}, are also combined with clustering methods for time series clustering and show certain merits. For example, edit distance with k-mediods \citep{elastic} and Kshape \citep{kshape} are shown to be strong baselines on UCR time series datasets \citep{ucrts}. 

\textbf{Shapelet-based Methods.} Other than analyzing the entire time series, shapelet-based time series clustering methods characterize time series with shapelets, which are subsequences that can discriminate time series of different classes. Specifically, U-shapelets \citep{ushape} discovers optimal shapelets by greedily searching high-quality subsequences that produce large $separation\ gap$s. Besides $separation\ gap$, other shapelet quality measurements are also used, such as Root-mean-square Standard Deviation, R-squared, and $I$ index \citep{compact}, which assess the deviation of clusters from different perspectives. Furthermore, FOTS-SUSH \citep{fots} develops a FOTS distance for shapelets discover, which is calculated based on the eigenvector decomposition and Frobenius correlation, to capture the complex relationships of time series under uncertainty. Different from discovering shapelets from time series subsequences, learning-based methods discover shapelets by objective optimization instead; for example, adopting a differentiable soft minimum distance for time series and shapelets \citep{lts}. AutoShape \citep{autoshape} integrates shapelet learning (as latent representation learning) with the framework of autoencoder, but it shows difficulty in learning understandable shapelets. Meanwhile, learning-based shapelets are mostly adopted for time series classification \citep{ussl,unlabeled,learning1} and normally require large prior knowledge of datasets \citep{learning2}.

\subsection{Semi-supervised Time Series Clustering}
Semi-supervised clustering methods improve the clustering accuracy by incorporating limited prior knowledge of the dataset. The prior knowledge normally is represented as a small set of constraints or labels \citep{cssc}. Constraint-based methods adjust the clustering process (e.g., COP-Kmeans \citep{COP-Kmeans}) to avoid violating the explicit constraints, i.e., whether two data points need to be in the same cluster (must-link) or different clusters (cannot-link) \citep{COP-Kmeans}. Existing methods develop different ways to utilize the constraints for time series clustering. SemiDTW \citep{semidtw} learns the optimal window size for the DTW distance, which violates the least number of manual must-link and cannot-link constraints. WSSNCut \citep{ssncut} uses slight supervision to integrate and weigh multiple different distance measurements of time series, and then adopts the semi-supervised normalized cut for clustering. COBRAS$^{TS}$ \citep{COBRAS} adopts the hierarchical division process and applies the manual constraints to refine the improperly divided groups. FssKmeans \citep{fsskmeans} first enriches the manual constraint set by propagating constraints through reverse nearest neighbours of time series, and then applies semi-supervised Kmeans for clustering. CDPS \citep{cdps} adopts the constraints to learn DTW-preserving shapelets, but that is different from our work as we aim at learning representative shapelets to discriminate time series of different classes.

Label-based methods adopt a small labeled subset, indicating the class information of time series, to supervise the clustering. Likewise, existing methods develop different strategies to utilize the labels for clustering. Seed-Kmeans \citep{seedkmeans} adopts the labeled subset to find optimal initialization of seed groups, considering that Kmeans are sensitive to the quality of initialization. CSSC \citep{cssc} further extends that and proposes a $compact\ degree$ to estimate the purity of clusters based on the labeled data contained. The labels are also used for density-based clustering, where SSDBSCAN \citep{ssdbscan} automatically finds a suitable density threshold by ensuring labeled data of different classes are not density-connected. 
Compared with constraints-based methods, label-based semi-supervised clustering methods can avoid the contradiction introduced by improper constraints and is also not sensitive to the order of labels \citep{cssc}. Following this, we utilize the labeled subset to discover representative shapelets for time series clustering.

\section{Preliminaries}
In the following content, a vector is denoted as a bold letter (e.g., $\boldsymbol{x}$), and a segment of $\boldsymbol{x}$ is denoted as $\boldsymbol{x}_{i:j}=\{x_{i},...,x_{j}\}$. A matrix is denoted as a capital letter (e.g., $M$), with the entry, the column, and the row denoted as $M_{i,j}$, $M_{:,j}$, and $M_{i,:}$, respectively.
A time series that contains $l$ real-valued numbers is denoted as $\boldsymbol{t}=\{t_{1},t_{2},...,t_{l}\}$. We preprocess a time series to be scale-invariant with z-normalization, which is denoted as $\boldsymbol{x}=\{x_{1},...,x_{l}\}$. 
A dataset is a collection of preprocessed time series and is denoted as $D=\{\boldsymbol{x}_{1},\boldsymbol{x}_{2},...,\boldsymbol{x}_{n}\}$, where $n$ is the size of $D$. The labels of time series in $D$ are denoted as $\boldsymbol{y}=\{y_{1},y_{2},...,y_{n}\}$ ($y_{i}\in R^{c})$, where $c$ is the number of classes.

\begin{figure}[!t]
\centering
\includegraphics[width=2.5in]{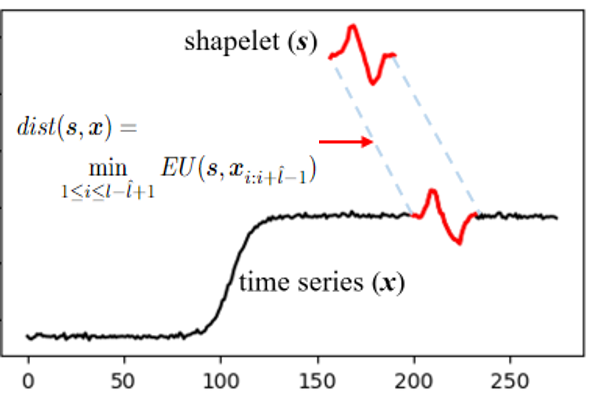}
\caption{The distance between shapelet $\boldsymbol{s}$ and time series $\boldsymbol{x}$ is the smallest Euclidean distance of $\boldsymbol{s}$ with a subsequence of $\boldsymbol{x}$.}
\label{fig:dist}
\end{figure}

A shapelet (denoted as $\boldsymbol{s}$) is a low-dimensional time series subsequence, i.e., $\boldsymbol{s}_{p}=\{s_{p1},...,s_{p\hat{l}}\}=\boldsymbol{x}_{p:p+\hat{l}-1}$ and $\hat{l}\ll l$. The distance of two low-dimensional shapelets, i.e., $\boldsymbol{s}_{1}$ and $\boldsymbol{s}_{2}$, is measured by Euclidean distance since they are insensitive to distortions \citep{ushape}:
\begin{equation}
EU(\boldsymbol{s}_{1},\boldsymbol{s}_{2})=\sqrt{\sum_{i=1}^{\hat{l}}(s_{1i}-s_{2i})^{2}}.
\label{eq:ed}
\end{equation}
The distance between a shapelet ($\boldsymbol{s}$) and a time series ($\boldsymbol{x}$) is defined as follows:
\begin{equation}
\begin{aligned}
dist&(\boldsymbol{s},\boldsymbol{x})= 
&\min_{1\leq i \leq l-\hat{l}+1} EU(\boldsymbol{s},\boldsymbol{x}_{i:i+\hat{l}-1}).
\end{aligned}
\label{eq:dist}
\end{equation}
We show the distance between a shapelet and a time series with the example in Fig. \ref{fig:dist}, using the Trace dataset \citep{ucrts}.

With a set of shapelets $\{\boldsymbol{s}_{1},\boldsymbol{s}_{2},...,\boldsymbol{s}_{k}\}$, a time series $\boldsymbol{x}$ is mapped to a low-dimensional distance-to-shapelets representation as follows:
\begin{equation}
\boldsymbol{h}=\{h_{1},h_{2}...,h_{k}\},
\label{eq:map}
\end{equation}
where $h_{i}=dist(\boldsymbol{s}_{i},\boldsymbol{x})$. The notations are summarized in Table \ref{tab:note} for convenience.

\begin{table}[!t]
\renewcommand{\arraystretch}{1.3}
\caption{Summary of key notations.}
\label{tab:note}
\centering
\begin{tabular}{cl}
\hline
Notation & Description \\
\hline
$n$ & the size of time series dataset $D$\\
$\hat{n}$ & the size of subset $D_{l}$ (labeled/pseudo-labeled) \\
$c$ & the number of classes in the dataset $D$\\
$\boldsymbol{x}$ & the time series\\
$l$ & the length of time series\\
$\hat{l}$ & the length of subsequences/shapelets\\
$\boldsymbol{s}_{p}$ & the time series subsequence, $\boldsymbol{s}_{p}=\boldsymbol{x}_{p:p+\hat{l}-1}$\\
$m$ & the number of subsequences of $\boldsymbol{x}$, $m=l-\hat{l}+1$\\
$\boldsymbol{h}$ & the distance-to-shapelets representation of $\boldsymbol{x}$\\
\hline
\end{tabular}
\end{table}

\section{The Proposed Method}
In this section, we first provide an overview of the SE-shapelets method, and then detail the two proposed techniques that discover representative shapelets for time series clustering.
\subsection{Method Overview}
\begin{figure*}[htbp]
\centering
\subfloat{\includegraphics[width=6.81in]{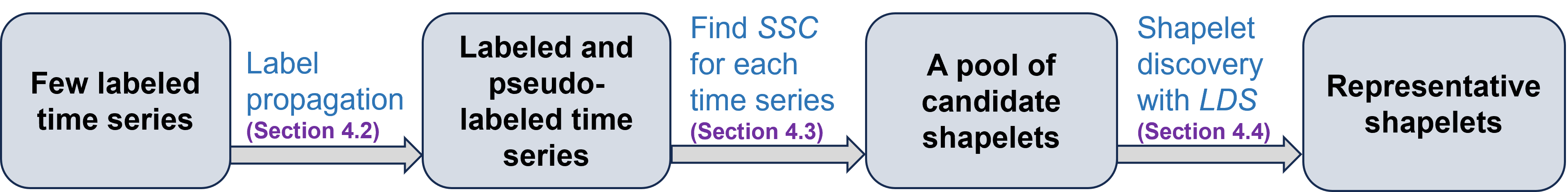}}
\caption{The flowchart of SE-shapelets to find representative shapelets. Then, discovered shapelets map original time series into the distance-to-shapelets representation with Eq. (\ref{eq:dist}) for the final clustering.}
\label{fig:flow}
\end{figure*}

The proposed SE-shapelets method first represents time series as low-dimensional distance-to-shapelet representations by Eq. (\ref{eq:dist}), with a set of discovered representative shapelets, and then adopts a clustering algorithm to cluster them. So, our aim is to discover shapelets with high representative abilities for time series clustering, using a small number of labeled time series. We adopt Spectral clustering \citep{spectral} for its effectiveness; other clustering algorithms, such as partition-based, density-based or model-based can also be used in SE-shapelets. Specifically, we discover shapelets from time series subsequences, considering that learning shapelets by objective optimization is prone overfitting for the small number of labeled/pseudo-labeled time series. The entire flow of shapelet discovery is shown in Fig. \ref{fig:flow}.

SE-shapelets first obtain some pseudo-labeled time series, by propagating labels of labeled time series to unlabeled time series, to enrich the labeling information. We then develop two techniques to discover representative shapelets (for time series clustering) by addressing the aforementioned problems of existing methods. First, we propose a \textit{salient subsequence chain} ($SSC$), which can extract salient subsequences (local features) of a time series. We only use salient subsequences extracted by $SSC$ as candidate shapelets, therefore pruning a large number of uninformative subsequences. Second, we propose a \textit{linear discriminant selection} ($LDS$) algorithm to discover optimal shapelets from the candidate shapelets, for time series clustering. With the labels and pseudo-labels, $LDS$ finds representative shapelets that can map time series of different classes into distinct and compact groups, which can easily be discovered by a clustering algorithm.

\subsection{Pseudo-label Propagation}
To enrich the limited labeling information, we first obtain some pseudo-labeled time series by label propagation. Following common practice, we adopt a simple propagation strategy, i.e., labeled time series propagate their labels to their nearest unlabeled time series  \citep{pseudo}. This strategy is based on the proven effectiveness of one nearest neighbour (1NN) classifier on various real-world time series datasets \citep{1nn}. Therefore, we also regard pseudo-labels acquired by unlabeled time series as confident labels and further propagate these pseudo-labels. The propagation is achieved by the following steps: 1) create a labeled set ($D_{l}$) that initially only contains the labeled time series; 2) for each time series in $D_{l}$, propagate their labels to their unlabeled nearest neighbours, and add the nearest neighbours with propagated pseudo-labels in $D_{l}$; 3) stop the propagation if no more nearest neighbour of time series in $D_{l}$ is unlabeled. During the process, if an unlabeled time series receives different labels from multiple labeled/pseudo-labeled time series, we choose the label of the time series that has the smallest distance.

\subsection{Salient Subsequence Extraction}
To avoid analyzing a large number of uninformative time series subsequences \citep{ushape}, we aim to extract several salient subsequences from every labeled/pseudo-labeled time series as shapelet candidates by finding a \textit{salient subsequence chain} ($SSC$). Inspired by the definition of ``salience'' in neuroscience (contrasts between items and their neighborhood) \citep{salience}, we regard salient subsequences as subsequences that are significantly different from their neighbouring subsequences. As an analogy, among the 2-character subsequences of ``000011111'', ``01'' is more salient than ``00'' or ``11''.

\begin{figure}[!t]
\centering
\includegraphics[width=3.3in]{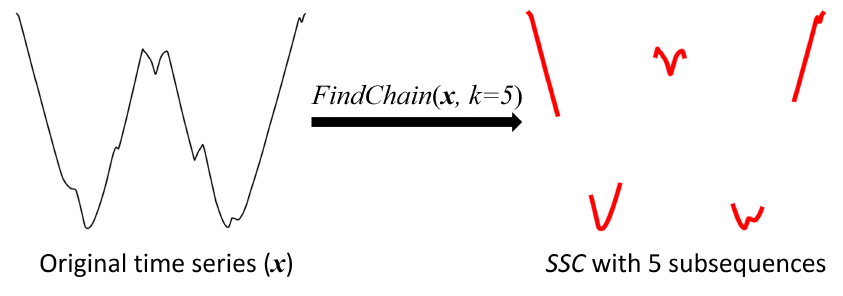}
\caption{$SSC$ of 5 subsequences for time series $\boldsymbol{x}$, extracted by $FindChain$.}
\label{fig:ssc}
\end{figure}

To explain $SSC$, here we assume that $SSC$ is used to extract $k$ salient subsequences from each labeled/pseudo-labeled time series. 
Before $SSC$, we first define the subsequence chain ($SC$) since $SSC$ is its special case.

\begin{definition}
A subsequence chain ($SC$) of time series $\boldsymbol{x}$ is a chain that contains $k$ timely-ordered and non-overlapping subsequences, that is, $SC=\{\boldsymbol{s}_{p_{1}},\boldsymbol{s}_{p_{2}},...,\boldsymbol{s}_{p_{k}}\}$, w.r.t. $p_{i}+\hat{l}\leq p_{i+1}, 1\leq i< k$.
\label{def:chain}
\end{definition}
Specifically, $SC$ expresses $k$ local features (subsequences) of $x$. Here $\boldsymbol{x}$ contains $m=l-\hat{l}+1$ subsequences that have lengths as $\hat{l}$, and any $k$ subsequences can form a unique $SC$. Specifically, we impose the subsequences in $SC$ to be timely ordered to represent their temporal relationships, and they are non-overlapping to preserve more information about the original time series.

We aim to select the $SC$ that best extracts salient local features of the time series. Similar to the ``salience'' in neuroscience, we describe the salience of a subsequence in a time series as follows:
\begin{definition}
\noindent The salience of $\boldsymbol{s}_{p_{i}}$ is measured by its difference from neighbouring subsequences in $SC$, i.e., $EU(\boldsymbol{s}_{p_{i-1}},\boldsymbol{s}_{p_{i}})$ $+EU(\boldsymbol{s}_{p_{i}},\boldsymbol{s}_{p_{i+1}})$. The greater the difference, the more salient the subsequence.
\label{def:salience}
\end{definition}
Note we only measure the difference with neighbouring subsequences rather than all subsequences because time series may contain recurrent local features, e.g., the two similar troughs in Fig. \ref{fig:ssc}.
Therefore, we define $sum(SC)$, which measures the total salience of subsequences in $SC$, as follows:
\begin{equation}
sum(SC)=\sum_{j=1}^{k-1}EU(\boldsymbol{s}_{p_{j}},\boldsymbol{s}_{p_{j+1}}).
\label{eq:scsum}
\end{equation}%

\begin{definition}
Salient subsequence chain ($SSC$) of $\boldsymbol{x}$ is the $SC$ that has the largest $sum(SC)$.
\label{def:psc}
\end{definition}
We show an example of the extracted $SSC$ that contains 5 subsequences in Fig. \ref{fig:ssc}, and it captures the five most salient local features of the time series.

An intuitive way to discover $SSC$ is the brute-force search from all possible $SC$s; however, this is infeasible in real use due to the large number of $SC$s. Hence, we propose an affordable $FindChain$ algorithm to find $SSC$ as follows:
\begin{itemize}
\item[--] we convert the $SSC$ discovery as the shortest path with $k$ nodes problem by designing a specific subsequence graph.
\item[--] we solve the shortest path with $k$ nodes problem using dynamic programming.
\end{itemize}

In $FindChain$, a directed acyclic subsequence graph, $G=(V,E)$, is constructed with nodes as the subsequences of $\boldsymbol{x}$. A directed edge ($W_{p,q}$) is added to $G$ from $s_{p}$ to $s_{q}$ if $p+\hat{l}\leq q$, with the weight as $EU(\boldsymbol{s}_{p},\boldsymbol{s}_{q})$, which ensures two connected subsequences are not overlapping. Then, a $SC=\{\boldsymbol{s}_{p_{1}},\boldsymbol{s}_{p_{2}},...,\boldsymbol{s}_{p_{k}}\}$ becomes a path (with $k$ nodes) on $G$, with $sum(SC)$ as the overall edge weights, and thus $SSC$ becomes the longest such path.
Note the first subsequence in $SSC$ do not necessarily need to be the beginning subsequence of $x$ ($\boldsymbol{s}_{1}=\boldsymbol{x}_{1:\hat{l}}$), and similarly for the last subsequence in $SSC$. Therefore, we further add two virtual nodes, $u$ and $v$, into $G$ for the convenience of $SSC$ discovery. 
$u$ is the virtual source and it has directed edges (with zero weights) to nodes of $\boldsymbol{s}_{p}, \forall 1\leq p\leq m$. $v$ is the virtual sink and has directed edges (with zero weights) from $\boldsymbol{s}_{p}, \forall 1\leq p\leq m$. 
Then, a $SC$ equals to a path containing $k+2$ nodes from $u$ to $v$ ($k$ subsequences and the virtual $u,v$) in $G$, and the problem of finding $SSC$ becomes finding the longest path among such paths. Since $G$ is directed and acyclic, finding the longest path can be solved as finding the shortest path by negating all the edge weights. The edge weights on $G$ are summarized as follows:
\begin{equation}
M_{p,q}=\begin{cases} 
      -EU(\boldsymbol{s}_{p-1},\boldsymbol{s}_{q-1}), &\text{if}\ 1 < q+\hat{l} \leq p < m+2 \\
      0, &\text{if}\ q = 1,p \neq 1 \\
      0, &\text{if}\ q \neq m+2,p = m+2 \\
      infinity, &\text{otherwise} \\
   \end{cases}
\label{eq:m}
\end{equation}
where $M_{p,q}=infinity$ indicates no edge.

The problem can be conveniently solved by dynamic programming. Specifically, a distance matrix, $A\in R^{(m+2)\times (k+2)}$, is used to store the accumulative distances of the so-far shortest paths. The row of $A$ represents the nodes in the topological order, i.e. $\{u,\boldsymbol{s}_{1},...,\boldsymbol{s}_{m},v\}$, and the column of $A$ is the size of the current $SSC$. The value of $A$ is calculated as follows:
\begin{equation}
A_{p,q}=\min\{A_{\alpha,q-1}+M_{\alpha,p}\ :\ 1\leq \alpha\leq m+2\},
\label{eq:a}
\end{equation}%
where $A_{p,q}$ is the accumulated distance of the shortest path from $u$ to $\boldsymbol{s}_{p}$, with $q$ nodes. We show the workflow of $FindChain$ and the extracted salient subsequences of an example time series in Fig. \ref{fig:ssc}.

The pseudo-code of $FindChain$ is shown in Algorithm \ref{alg:1}. At lines 1-2, $M$ and $A$ are initialized. $M$ is updated with the negative edge weights on the graph at lines 4-6. The dynamic programming process that finds the shortest path is shown at lines 7-11, and the $SSC$ is discovered for $\boldsymbol{x}$ at lines 12-13.
For simplicity, we discover $k$ shapelets from these extracted salient subsequences of labeled/pseudo-labeled time series for the clustering, i.e., representing a time series with $k$ shapelets. The time complexity of $FindChain$ is $O(km^{2})$ and space complexity $O(m^{2}+km)$, where $k$ is the size of $SSC$ and $m$ is the number of nodes/subsequences in $G$.

\begin{algorithm}[tb]
\caption{FindChain}
\label{alg:1}
\textbf{Input}: Time Series $\boldsymbol{x}$, subsequence length $\hat{l}$, chain size $k$

\begin{algorithmic}[1] 
\State Initialize $M$=matrix of $(m+2)\times (m+2)$ infinities. \\
 \ \ \ \ \ \ \  \ \ \ \ $A$=matrix of $(m+2)\times (k+2)$ zeros.
 
\State $\{\boldsymbol{s}_{1},...,\boldsymbol{s}_{m}\}=\{\boldsymbol{x}_{1:1+\hat{l}},...,\boldsymbol{x}_{m:m+\ell}\}$
\For{each $(\boldsymbol{s}_{p},\boldsymbol{s}_{q}, p+\hat{l}\leq q)$ pair}
\State $M_{p+1,q+1}=-EU(\boldsymbol{s}_{p},\boldsymbol{s}_{q})$.
\EndFor
\For{each $2\leq q\leq k+2$}
\For{each $1\leq p\leq m+2$}
\State $A_{p,q}=\min\{A_{\alpha,q-1}+M_{\alpha, p}\ :\ \leq \alpha\leq m\}$.
\EndFor
\EndFor
\State Retrieve the $ShortestPath$ starting from $A_{m+2,k+2}$.
\State $SSC$=$ShortestPath\setminus \{u,v\}$.
\State \textbf{return} The salient subsequence chain $SSC$.
\end{algorithmic}
\end{algorithm}

\subsection{Shapelet Discovery}
After obtaining $k\hat{n}$ salient subsequences extracted by applying $FindChain$ on $\hat{n}$ labeled/pseudo-labeled time series, we aim to select $k$ best subsequences as shapelets for the clustering. So, we formulate the shapelet discovery as finding $k$ optimal shapelets from a pool of shapelet candidates. To ensure the discovered shapelets are representative, we first adopt Kmeans to these subsequences and discover $\gamma$ ($=\beta k$, $\beta$ is a positive integer and we set it as 2 based on the experiments) clusters. The centers of the clusters are regarded as the final candidate shapelets.

\begin{figure}[!t]
\centering
\includegraphics[width=3.2in]{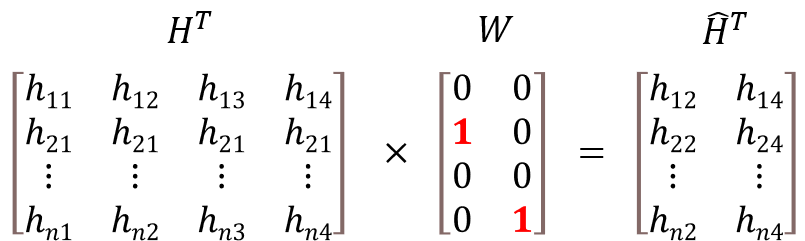}
\caption{An example of the selection matrix $W$, and it selects the second and the fourth columns of $H$ to obtain $\hat{H}$.}
\label{fig:w}
\end{figure}

Among the $\gamma$ candidate shapelets, i.e., $\{\boldsymbol{s}_{1},\boldsymbol{s}_{2},...,\boldsymbol{s}_{\gamma}\}$, we adopt a shapelet selection matrix to choose $k$ candidate shapelets as shapelets, which are denoted as $\{\boldsymbol{s}_{p_{1}},\boldsymbol{s}_{p_{2}},...,\boldsymbol{s}_{p_{k}}\}$ and $p_{i}\in \{1,2,...,\gamma\}$. Specifically, with the candidate shapelets, the labeled/pseudo-labeled time series are mapped to the distance-to-shapelets representation $H\in R^{\gamma\times \hat{n}}$, where $H_{:,i}=\boldsymbol{h}_{i}$ is calculated with Eq. (\ref{eq:dist}). In contrast, we denote the representation regarding the $k$ shapelets (waiting to be discovered) as $\hat{H}\in R^{k\times \hat{n}}$. With $H$ and $\hat{H}$, we adopt a binary shapelet selection matrix ($W\in R^{\gamma\times k}$) and formulate the shapelet discovery as follows:
\begin{equation}
\hat{H}=W^{T}H.
\label{eq:mapping}
\end{equation}%
The row and column of $W$ correspond to the size of candidate shapelets ($\gamma$) and the number of expected shapelets ($k$), respectively. We show an example of $W$ that selects two shapelets from four candidate shapelets ($W\in R^{4\times 2}$) in Fig. \ref{fig:w}.
For shapelet selection, the binary $W$ has the following three characteristics: 
\begin{itemize}
\item[--]  $\sum_{p,q}W_{p,q}=k$. $W$ has $k$ non-zero entries ($=1$), and each non-zero entry selects a corresponding candidate shapelet as a shapelet. 

\item[--]  For each column $q$, $\sum_{p}W_{p,q}=1$. This non-zero entry in the column selects a shapelet; for example, in Fig. \ref{fig:w}, $\boldsymbol{s}_{1}$ is selected as a shapelet as $W_{1,1}=1$.

\item[--]  For each row $p$, $\sum_{q}W_{p,q}\leq 1$. Each row of $W$ has at most one non-zero entry to ensure each candidate shapelet can only be selected once. 
\end{itemize}
Therefore, determining $W$ is equivalent to determining the row indices of its non-zero entries.
We denote the row indices of non-zeros entries in $W$ as $\{p_{1},p_{2},...,p_{k}\}$, where $p_{i}!=p_{j}$ for any $(p_{i}, p_{j}), i!=j$. Thus, the values of entries in $W$ are:
\begin{equation}
W_{p,q}=\begin{cases} 
      1, &W_{p,q}\in \{W_{p_{1},1},W_{p_{2},2},...,W_{p_{k},k}\}\\
      0, &\text{otherwise}. \\
   \end{cases}
\end{equation}
The objective is to find the optimal $W$ (or the optimal row indices of non-zero entries) to select the best shapelets.

To find the optimal $W$, we propose a \textit{linear discriminant selection} ($LDS$) method, which integrates the binary shapelet selection matrix ($W$) with the linear discriminant analysis \citep{lda}, to select the $k$ best shapelets from the $\gamma$ candidate shapelets. The aim is to select shapelets that can map time series into distinctive groups for convenient clustering; that is, the groups are expected to be compact and far away from each other.
With the binary shapelet selection matrix, the objective function of $LDS$ is defined as follows:
\begin{equation}
\ell=\max_{W}\frac{tr(W^{T}S_{B}W)}{tr(W^{T}S_{W}W)},
\label{eq:lda}
\end{equation}%
where $S_{B}\in R ^{\gamma\times \gamma}$ reflects distances among different groups and $S_{W}\in R ^{\gamma\times \gamma}$ measures the compactness of groups. Specifically, $S_{B}$ is the between-class scatter matrix defined by the covariance of the class means:
\begin{equation}
S_{B}=\sum_{i=1}^{c}|\boldsymbol{c}_{i}|(\boldsymbol{u}_{i}-\overline{\boldsymbol{u}})(\boldsymbol{u}_{i}-\overline{\boldsymbol{u}})^{T},
\label{eq:sb}
\end{equation}%
where $\boldsymbol{u}_{i}=\frac{1}{|\boldsymbol{c}_{i}|}\sum_{\boldsymbol{x}_{j}\in \boldsymbol{c}_{i}}H_{:,j}$ and $\boldsymbol{u}=\frac{1}{\hat{n}}\sum_{j=1}^{\hat{n}}H_{:,j}$. 
$S_{W}$ is the within-class scatter matrix defined as follows:
\begin{equation}
S_{W}=\sum_{i=1}^{c}\sum_{\boldsymbol{x}_{j}\in \boldsymbol{c}_{i}}(H_{:,j}-\boldsymbol{u}_{i})(H_{:,j}-\boldsymbol{u}_{i})^{T}.
\label{eq:sw}
\end{equation}%
Directly seeking an analytical solution of optimal $W$ for Eq. (\ref{eq:lda}) is challenging because of the binary nature of $W$ and its previously defined characteristics. Therefore, without loss of generality, we rewrite Eq. (\ref{eq:lda}) as:
\begin{equation}
\begin{split}
\ell&=\max_{W}(tr(W^{T}S_{B}W)-\lambda tr(W^{T}S_{W}W))\\
&=\max_{W}tr(W^{T}(S_{B}-\lambda S_{W})W)\\
&=\max_{W}tr(\Theta \Gamma),
\end{split}
\label{eq:lda1}
\end{equation}%
where $\lambda$ is a constant weight, $\Theta=WW^{T}$ and $\Gamma=S_{B}-\lambda S_{W}$.
Because $W$ only has $k$ non-zeros entries ($=1$, as $\{W_{p_{1},1},W_{p_{2},2},...,W_{p_{k},k}\}$) and they do not locate in the same row or column, $\Theta=WW^{T}$ ($\in R^{\gamma\times \gamma}$) is a square matrix that only has $k$ non-zero entries on the diagonal:
\begin{equation}
\Theta_{p,q}=\begin{cases} 
      1, &p=q\in  \{p_{1},p_{2},...,p_{k}\} \\
      0, &\text{otherwise}, \\
   \end{cases}
\end{equation}
Therefore, we have:
\begin{equation}
tr(\Theta \Gamma)=\sum_{i=1}^{k}\Gamma_{p_{i},p_{i}}. 
\label{eq:sw}
\end{equation}%
Based on Eq. (\ref{eq:sw}), the analytical solution of $W$ that maximizes Eq. (\ref{eq:lda1}), i.e., the optimal $\{p_{1},p_{2},...,p_{k}\}$, becomes the row indices of the $k$ largest diagonal entries of $\Gamma$. The time complexity of $LDS$ is $O(c\gamma^{2}\hat{n})$ and the space complexity is $O(\gamma^{2}+\gamma k)$.

With $\{p_{1},p_{2},...,p_{k}\}$, the indices of non-zeros entries within the optimal $W$, we discover shapelets as $\{\boldsymbol{s}_{p_{1}},\boldsymbol{s}_{p_{2}},...,\boldsymbol{s}_{p_{k}}\}$, where $\boldsymbol{s}_{p_{i}}\in\{\boldsymbol{s}_{1},\boldsymbol{s}_{2},...,\boldsymbol{s}_{\gamma}\}$. Then, we map each $\boldsymbol{x}_{i}\in D$ to its distance-to-shapelets representation ($\boldsymbol{h}_{i}$) with Eq. (\ref{eq:map}), and adopt Spectral clustering on the new distance-to-shapelets representation of all time series to obtain the final clusters.

Although the proposed SE-shapelets primarily targets univariate time series clustering, it is convenient to extend SE-shapelets to multivariate time series. A straight forward way is to discover an SSC for each variable in a multivariate time series sample. Parallel computing can be adopted to overcome the increased computational overhead. This simple extension assumes variables are independent and may miss the complex correlation patterns among them. Therefore, other operations such as constraints and heuristics are expected to be further developed for more effective shapelet discovery, which ensures discovered shapelets capture proper correlation patterns for multivariate time series.

\section{Evaluation}
In this section, we evaluate and compare the proposed SE-shapelets algorithm with baseline and most recent semi-supervised time series clustering methods. We run all of the experiments on a Linux platform with 2.6GHz CPU and 132GB RAM. 
The source code of SE-shapelets and the detailed results are available in our GitHub repository\footnote{https://github.com/brcai/SE-shapelets}.

\subsection{Datasets}
We use 85 time series datasets from the UCR time series archive \citep{ucrts} for evaluation. The sizes of these datasets range from 40 to 16,637, and time series lengths range from 24 to 2,709; we refer our GitHub repository for the detailed statistics of the datasets. Each dataset has a training set and a testing set (both have labels), and we use them both for clustering. The clustering results of evaluated methods are compared with the ground truth labels for the accuracy measurement.

\begin{figure*}[htbp]
\centering
\subfloat{\includegraphics[width=1.61in]{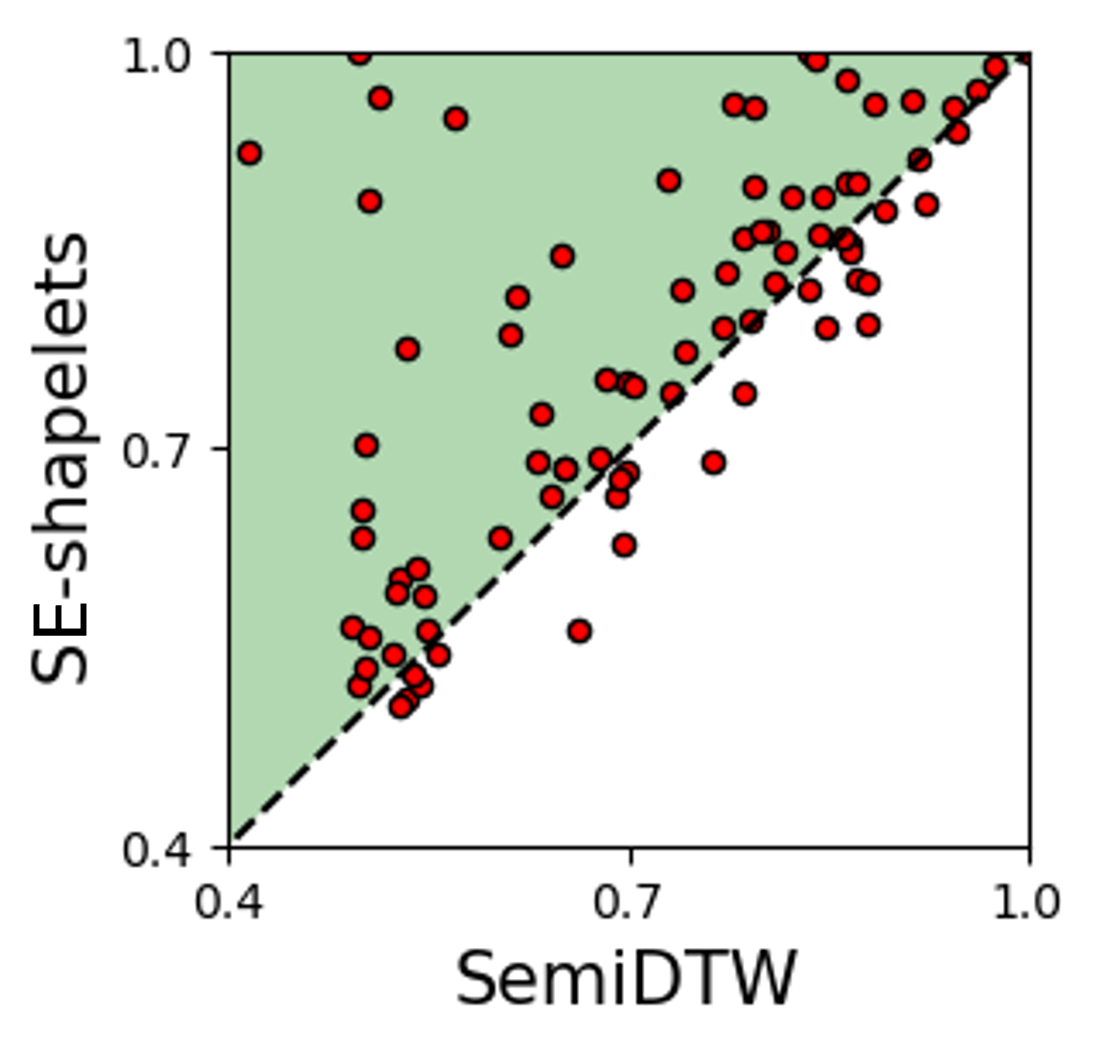}}\hspace{0.07in}
\subfloat{\includegraphics[width=1.61in]{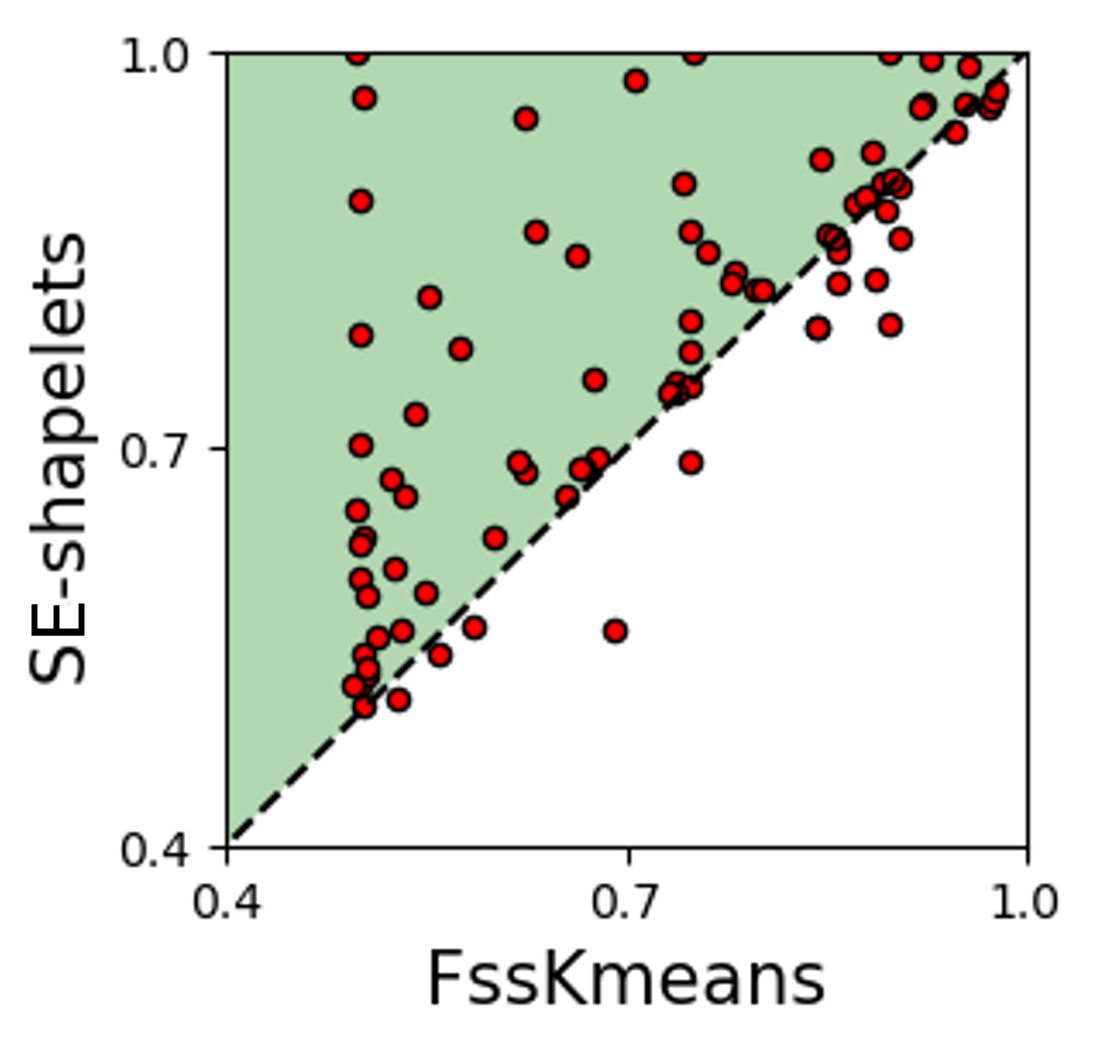}}\hspace{0.07in}
\subfloat{\includegraphics[width=1.61in]{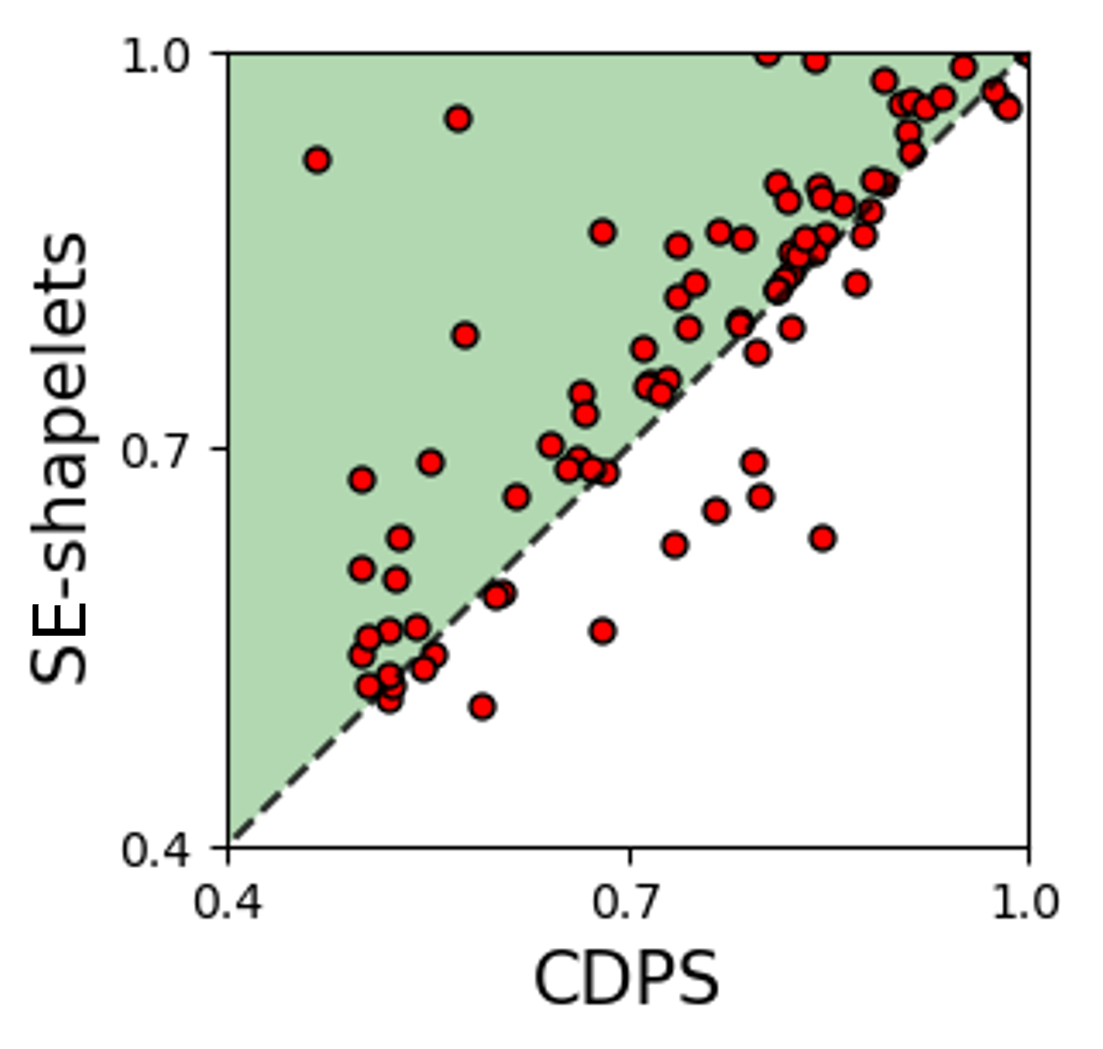}}\hspace{0.07in}
\subfloat{\includegraphics[width=1.61in]{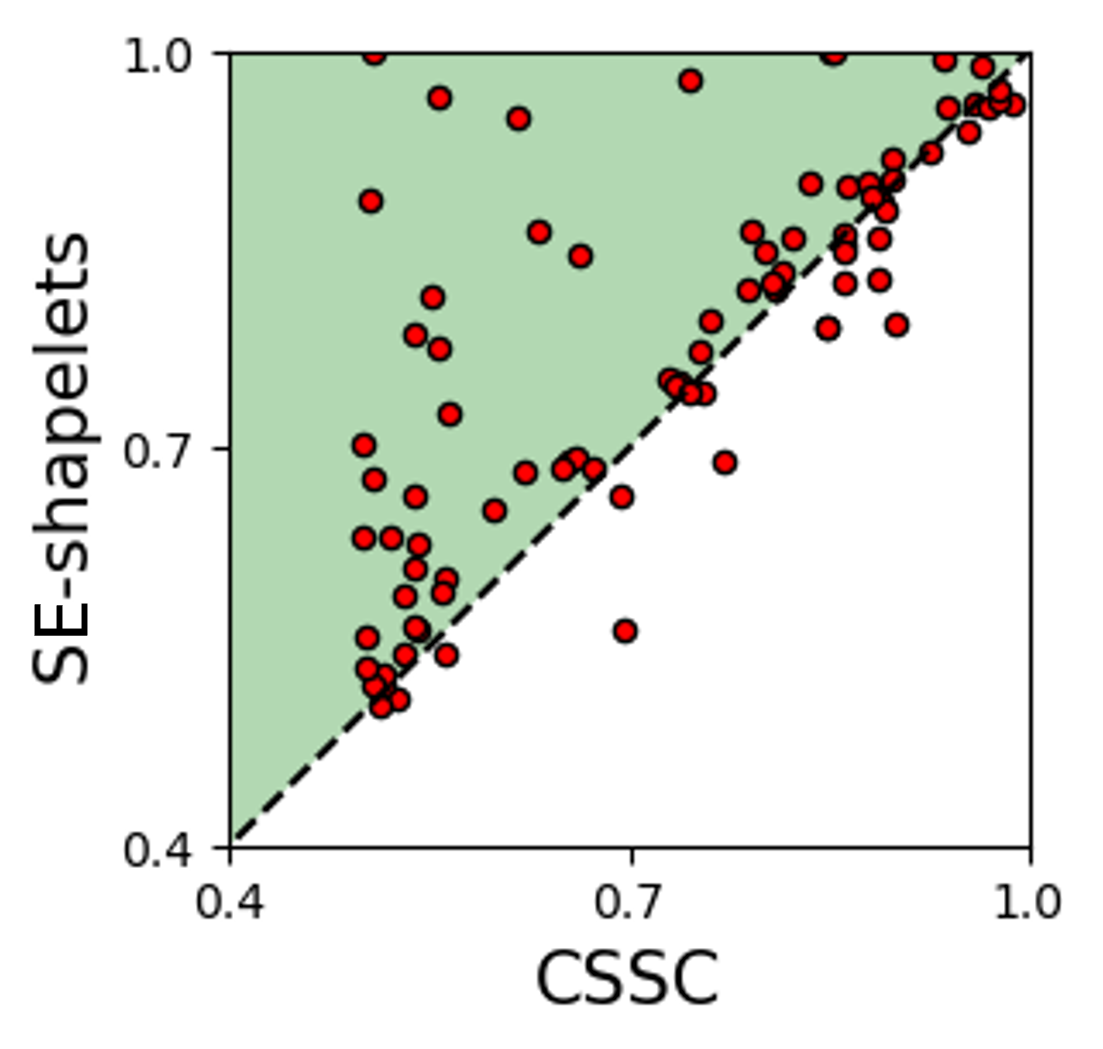}}
\caption{The comparisons of SE-shapelets with SemiDTW, FssKmeans, CDPS and CSSC, with respect to RI on the UCR time series datasets. Each dot represents a dataset, and SE-shapelets achieves better RI than the compared method if the dot locates in the shaded area.}
\label{fig:scatter}
\end{figure*}

\subsection{Performance Metric}
The clustering accuracy is measured by Rand Index following \citep{fsskmeans,semidtw}. Rand Index penalizes false positive and false negative clustering results and is defined as follows:
\begin{equation}
\text{RI} = \frac{TP+TN}{TP+TN+FP+FN}, 
\label{eq:eu}
\end{equation}%
where $TP$ (true positive) is the number of correctly clustered time series pairs; $TN$ is the number of correctly separated pairs; $FP$ means the number of pairs that are wrongly clustered; $FN$ is the number of time series pairs that are wrongly separated in different clusters. RI$\in (0,1]$, and a higher RI indicates a better performance.

\subsection{Baseline Methods}
We choose two types of counterpart semi-supervised time series clustering methods to compare with the proposed SE-shapelets method, i.e., constraint-based methods and label-based methods. For constraint-based methods, we choose SemiDTW \citep{semidtw}, which adopts the manual constraints to determine the optimal warping window for DTW distance, FssKmeans \citep{fsskmeans} that enriches the provided constraints with a constraint propagation heuristic, and CDPS \citep{cdps} that learns DTW-preserving shapelets. For label-based methods, we choose Seeded-Kmeans \citep{seedkmeans}, which finds optimal initialization with labeled time series, and the recent CSSC \citep{cssc}, which evaluates the compactness of clusters with labeled time series.
These methods are briefed as follows:
\begin{itemize}
\item[--] \textbf{SemiDTW} \citep{semidtw} is a semi-supervised time series clustering method based on DTW distance and density-based clustering, and it uses a small number of must/cannot-link constraints to discover optimal DTW window size.
\item[--] \textbf{FssKmeans} \citep{fsskmeans} extends the manual must/cannot-link constraints by propagating constraints to reverse nearest neighbours of constrained time series. The extended and original constraints are adopted on semi-supervised Kmeans for time series clustering. 
\item[--] \textbf{Seeded-Kmeans} \citep{seedkmeans} is a label-based semi-supervised clustering method, and it uses labels to determine optimal seeds, which produces high-quality initialization, for Kmeans.
\item[--] \textbf{CSSC} \citep{cssc} introduces a compact degree of clusters (measured by labels appearing in clusters) to assess the qualities of these clusters. The compact degree is jointly optimized with the clustering objective to discover compact clusters.
\item[--] \textbf{CDPS} \citep{cdps} learns constrained DTW-preserving shapelets to overcome time series distortions by approximating DTW. The process is guided by must/cannot-link constraints.
\end{itemize}

To better demonstrate the effectiveness of the semi-supervised discovery of shapelets in SE-shapelets, we further include the unsupervised shapelet-based clustering method, U-shapelets \citep{ushape}, for the discussion.

\subsection{Experiment Setup}
Other than SemiDTW that provides source code, we implement SE-shapelets and other compared methods with Python 3.7. For SemiDTW and CDPS, we use the recommended parameter setup to obtain clustering results \citep{semidtw,cdps}, and provide FssKmeans, Seeded-Kmeans and CSSC the cluster number for clustering. We set the trade-of parameter of CSSC as 0.7 (for cluster compactness and clustering objective), following \citep{cssc}.
The optimal parameters of SE-shapelets and U-shapelets are searched by applying grid search on the labeled and pseudo-labeled time series.
Specifically, for U-shapelets, we search the optimal shapelet number from $\{2,3,...,9\}$, the optimal shapelet length from $\{\frac{l}{30},\frac{l}{25},...,\frac{l}{10}\}$, where $l$ is the length of time series. For SE-shapelets, in addition to the two parameters searched the same as U-shapelets, we search the optimal $\lambda$ from $\{0.1,1,10\}$. In the Spectral clustering used by SE-shapelets, we simply choose the \textit{rbf} kernel (gamma=1.0) for all experiments.

The \textit{\textbf{level of semi-supervision}} \citep{cdps} is fixed as $5\%$, i.e., the number of labels or constraints divided by the dataset size, considering that it is difficult to acquire labels in real-life. That means, we randomly select labeled time series (for SE-shapelets, Seeded-Kmeans and CSSC) and generate must-link/cannot-link constraints (for SemiDTW, FssKmeans and CDPS) that cover all time series classes.

\begin{table*}[h]
\centering
\caption{Clustering accuracy comparison of SE-shapelets and semi-supervised counterpart methods (with the best in bold).}
\begin{tabular}{lcccccc}
\hline 
& SE-shapelets                          & CDPS  & SemiDTW   &FssKmeans  &Seeded-Kmeans  &CSSC    \\
\hline
Avg. RI	  &\textbf{0.778}	        &0.741  &0.713 	    &0.711      &0.685  &0.722        \\ 
Avg. Rank  &\textbf{1.1}	            &2.4    &2.4        &2.5        &3.8    &1.9          \\ 
$1^{st}$ Rank (datasets) &\textbf{45}   &14     &14         &6          & 1     & 11          \\ 
\hline
\end{tabular}
\label{tab:semisuper}
\end{table*}

\subsection{Main Results}
\begin{figure}[!htbp]
\centering
\subfloat{\includegraphics[width=3.3in]{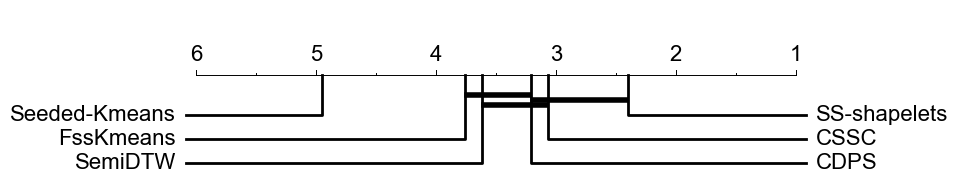}}
\caption{Ranks (lower is better) of the compared methods on the UCR datasets. The solid lines connect the methods that are not significantly different using Holm-Bonferroni method.}
\label{fig:cd}
\end{figure}

In this experiment, we compare SE-shapelets with the counterpart semi-supervised time series clustering methods to show its effectiveness. We summarize clustering accuracy (measured by RI) results on UCR time series datasets in Table \ref{tab:semisuper} (detailed in Appendix). SE-shapelets achieves the highest average clustering accuracy (0.778) and the largest average rank (1.1) among all the semi-supervised methods. Compared with CSSC, which obtains the second-highest average rank (1.9), SE-shapelets improved RI by around $7.8\%$, and improved RI of Seeded-Kmeans (0.685, the lowest) by $13.5\%$. Although all adopt constraints for the semi-supervision, CDPS and SemiDTW (average RI are 0.741 and 0.713, respectively) perform better than FssKmeans (average RI is 0.711) partly due to the advantage of distortion-invariant DTW distance, but SE-shapelets still outperforms them both.  
In addition, SE-shapelets achieves the best clustering accuracy on most datasets (45), which is significantly better than the second-best methods CDPS and SemiDTW (14). 
In addition, we show the statistical comparisons of the above methods with a critical difference diagram in Fig. \ref{fig:cd}. The hypothesis that not all of these methods are significantly different is rejected by Holm-Bonferroni method \citep{cd}, and SE-shapelets has the highest Wilcoxon test ranking; that shows SE-shapelets achieves a statistically significant improvement over the compared methods.

\begin{figure}[!htbp]
\centering
\subfloat{\includegraphics[width=2.53in]{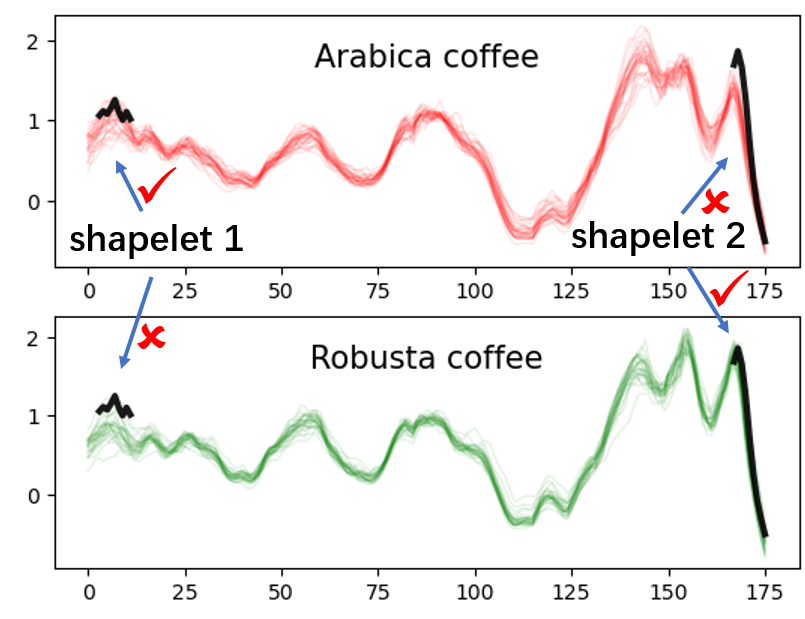}}
\caption{Two shapelets discovered by SE-shapelets (in red color) to discriminate the two types of coffees (Arabica and Robusta) in the Coffee dataset. Shapelet 1 and shapelet 2 capture representative local features of Arabica coffee and Robusta coffee, respectively.}
\label{fig:coffee}
\end{figure}

We also show the pair-wise comparison of SE-shapelets with counterpart methods as scatter plots in Fig. \ref{fig:scatter}. We observe that the clustering accuracy of SE-shapelets is better than SemiDTW, FssKmeans, CDPS, and CSSC on most datasets. Especially, SE-shapelets outperforms FssKmeans on 65 datasets; while on the rest 20 datasets, the accuracy of SE-shapelets is only slightly lower than that of FssKmeans. These results demonstrate the effectiveness of SE-shapelets that discovers shapelets for time series clustering.

We use an example (Coffee) dataset to show the shapelets discovered by SE-shapelets for the clustering. Coffee dataset contains the spectrographs of two types of coffees (Arabica and Robusta) as shown in Fig. \ref{fig:coffee}, respectively. The spectrographs of Arabica and Robusta are generally similar, but SE-shapelets captures two local differences to discriminate them. That is, the tiny spike on the left of Arabica's spectrographs and the great drop near the end of Robusta's spectrographs. Hence, with these two representative shapelets, SE-shapelets accurately clusters the two types of coffees (RI = 1.0).

\begin{figure}[!htbp]
\centering
\subfloat{\includegraphics[width=1.59in]{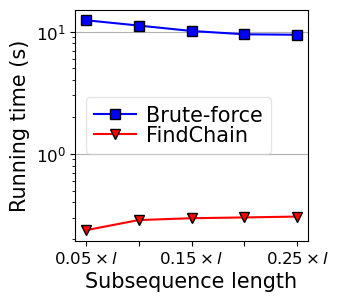}}\hspace{0.1in}
\subfloat{\includegraphics[width=1.56in]{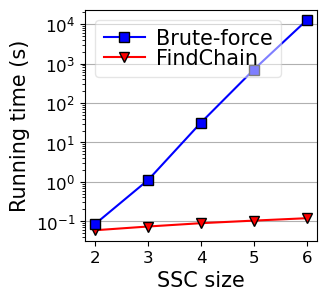}}
\caption{Running time of $FindChain$ and the brute-force search to discover $SSC$ for a time series randomly picked from StarLightCurves dataset.}
\label{fig:findchain}
\end{figure}

\begin{figure*}[htbp]
\centering
\subfloat[BeetleFly (SE-Shapelets)]{\includegraphics[width=1.63in]{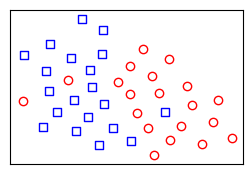}}\hspace{0.06in}
\subfloat[Coffee (SE-shapelets)]{\includegraphics[width=1.63in]{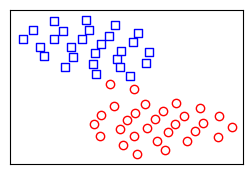}}\hspace{0.06in}
\subfloat[CBF (SE-shapelets)]{\includegraphics[width=1.63in]{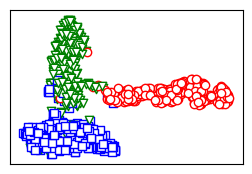}}\hspace{0.05in}
\subfloat[TwoLeadECG (SE-shapelets)]{\includegraphics[width=1.63in]{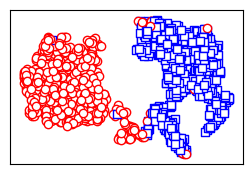}}\\
\subfloat[BeetleFly (U-shapelets)]{\includegraphics[width=1.63in]{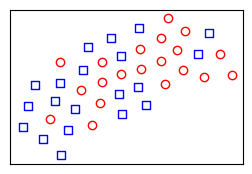}}\hspace{0.05in}
\subfloat[Coffee (U-shapelets)]{\includegraphics[width=1.63in]{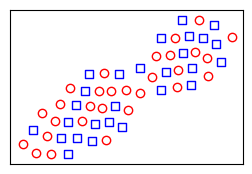}}\hspace{0.05in}
\subfloat[CBF (U-shapelets)]{\includegraphics[width=1.63in]{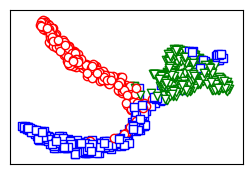}}\hspace{0.05in}
\subfloat[TwoLeadECG (U-shapelets)]{\includegraphics[width=1.63in]{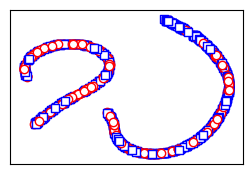}}
\caption{The visualization of the distances among distance-to-shapelets representations of time series, with shapelets discovered by SE-shapelets and U-shapelets, respectively. Each symbol represents a time series, and time series of different classes are represented by symbols of different shapes/colors.}
\label{fig:tsne}
\end{figure*}

\begin{figure}[htbp]
\centering
\subfloat{\includegraphics[width=3.3in]{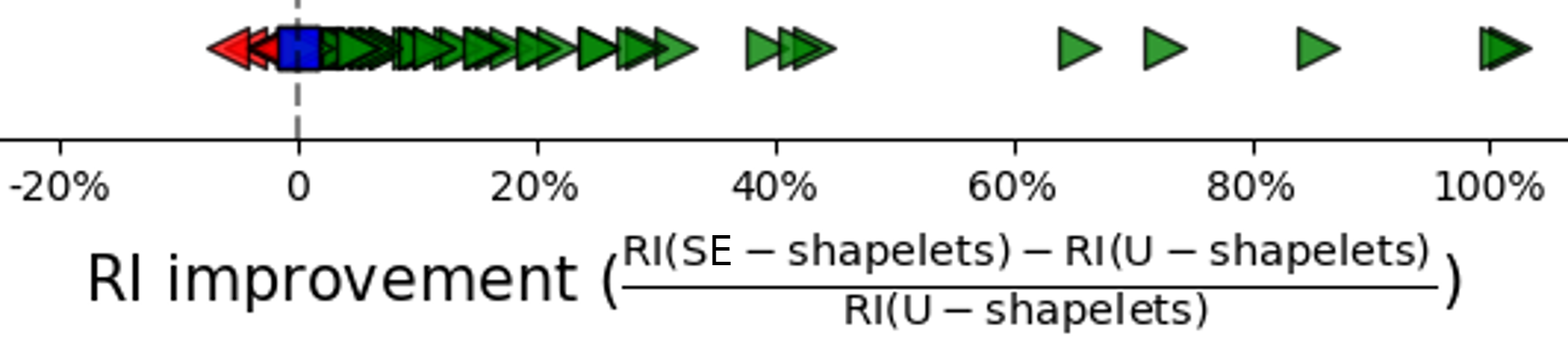}}\\
\caption{The distribution of clustering accuracy (RI) improvement (SE-shapelets improves the clustering accuracy of U-shapelets). Each symbol represents a time series dataset.}
\label{fig:distribution}
\end{figure}

\subsection{Ablation Analysis}
To understand the effectiveness of the two proposed techniques ($SSC$ that extracts salient subsequences and $LDS$ that select representative shapelets using the labels/pseudo-labels) in SE-shapelets, we develop three variants for the comparison and the results are shown in Table \ref{tab:ablation}. In addition, we also include the comparison with directly applying Spectral clustering for time series clustering (w/o shapelets). 
\begin{table}[h]
\centering
  \caption{Ablation analysis. <, = and > indicate the number of datasets that have \textbf{worse}, \textbf{equal} or \textbf{better} clustering accuracy than \textbf{SE-shapelets}, respectively.}
  \begin{tabular}{lcccc}
  	\hline & $<$ & $=$ & $>$ & Average RI {\small (decrease)}\\
    \hline
    w/o $SSC$       &   75&	  0&	  10&   0.725 {\small (6.8\%)}  \\
    w/o $LDS$       &   66&	  6&	  13&   0.737 {\small (5.2\%)}  \\
    w/o $LDS$*      &   82&	  1&	   2&   0.700 {\small (10.0\%)}  \\
    w/o shapelets   &   80&   0&       5&   0.541 {\small (30.4\%)}  \\
    \hline
  \end{tabular}
\label{tab:ablation}
\end{table}    

The average clustering accuracy of SE-shapelets is reduced by 6.8\%, when $SSC$ is removed from SE-shapelets (w/o $SSC$), and the clustering accuracy decreases in 66 (out of 85) datasets, because shapelets are selected from pools that contain many uninformative candidates, rather than explicitly extract salient subsequences (by $SSC$ of SE-shapelets).
In addition, we show the efficiency of the proposed $FindChain$ algorithm, which discovers $SSC$ for an input time series, with the brute-force search algorithm. We randomly select one time series ($l=1024$) from the largest StarLightCurves dataset among the UCR time series dataset (considering both dataset size and time series length), and run $FindChain$ and the brute-force search with different subsequence lengths and $SSC$ sizes.
The results in Fig. \ref{fig:findchain} (a) show that the running time of $FindChain$ slightly increases with larger subsequence length, due to the more edge weight calculations on the subsequence graph in Eq. (\ref{eq:m}); but the running time is still close to 0 second with the largest subsequence length ($0.25\times l$). On the contrary, the running time of the brute-force search gradually decreases with larger subsequence length, i.e., from around 12 seconds to 10 seconds, due to the smaller number of subsequences; but the running time is still more than one magnitude larger than that of $FindChain$. Meanwhile, the results for $SSC$ size (Fig. \ref{fig:findchain} (b)) again shows that the running time of $FindChain$ slightly increases but is still close to 0 second when discovering 6 salient subsequences; however, the brute-force search requires more than $1\times 10^{4}$ seconds to find such a $SSC$.

We further develop two variants that replace the proposed $LDS$ with $information\ gain$ \citep{gain0} (w/o $LDS$) and $nearest\ neighbour\ accuracy$ \citep{sd} (w/o $LDS*$) in SE-shapelets, respectively. As the results in Table \ref{tab:ablation} show, they reduce the average accuracy of SE-shapelets by 5.2\% and 10.0\%, respectively, and also decrease clustering accuracy in most datasets (66 and 82). That demonstrates that the proposed $LDS$ of SE-shapelets is more suitable for the clustering objective and can effectively select representative shapelets. 

Finally, we remove the shapelets from SE-shapelets and only use Spectral clustering for time series clustering (w/o shapelets). We can see that the clustering accuracy is reduced in 80 datasets, with the average clustering accuracy reduced by $30.4\%$; that shows the representative shapelets discovered by SE-shapelets significantly improves the performance of time series clustering.

\subsection{SE-shapelets vs U-shapelets}
To understand the effectiveness of the semi-supervised strategy of SE-shapelets, we further compare the performance of SE-shapelets with the U-shapelets (unsupervised shapelet-based clustering), and the results are summarized in Table \ref{tab:ushape}. With slight supervision, SE-shapelets improves the average clustering accuracy of U-shapelets by $10.8\%$, from 0.702 to 0.778. Among the 85 datasets, SE-shapelets achieves higher clustering accuracy in 68 datasets. 
\begin{table}[h]
\centering
\caption{Accuracy comparison between SE-shapelets and U-shapelets (with the best in bold).}
\begin{tabular}{lcc}
\hline 
& SE-shapelets & U-shapelets \\
\hline
Avg. RI	  &\textbf{0.778}	&0.702 \\ 
Higher RI {\small (datasets)} &\textbf{68}	    &14   \\ 
\hline
\end{tabular}
\label{tab:ushape}
\end{table}

\begin{figure*}[htbp]
\centering
\subfloat[BeetleFly]{\includegraphics[width=1.63in]{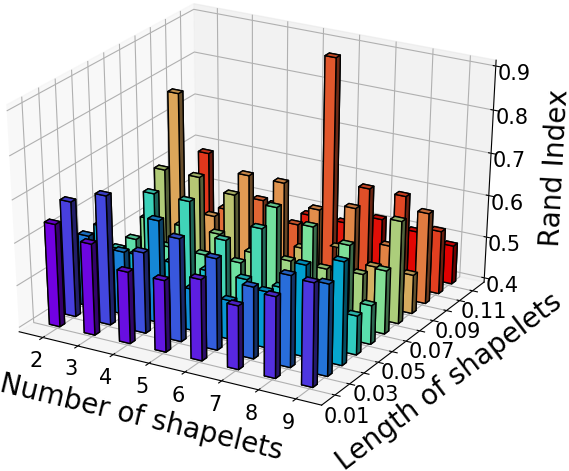}}\hspace{0.05in}
\subfloat[Coffee]{\includegraphics[width=1.63in]{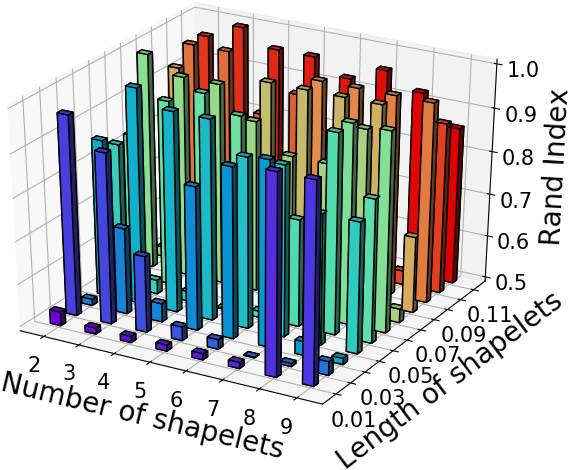}}\hspace{0.05in}
\subfloat[CBF]{\includegraphics[width=1.63in]{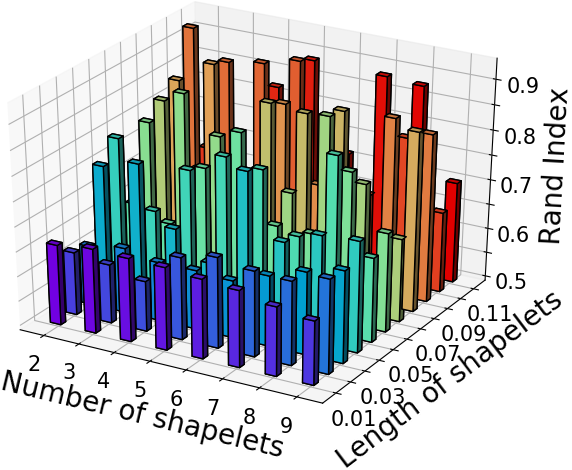}}\hspace{0.05in}
\subfloat[TwoLeadECG]{\includegraphics[width=1.63in]{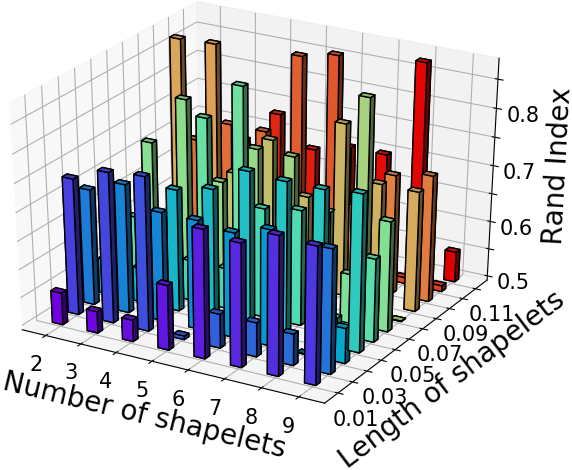}}
\caption{Parameter analysis of SE-shapelets, i.e., the number of shapelets and the length of shapelets, with respect to clustering accuracy.}
\label{fig:param}
\end{figure*}

\begin{figure*}[htbp]
\centering
\subfloat[BeetleFly]{\includegraphics[width=1.63in]{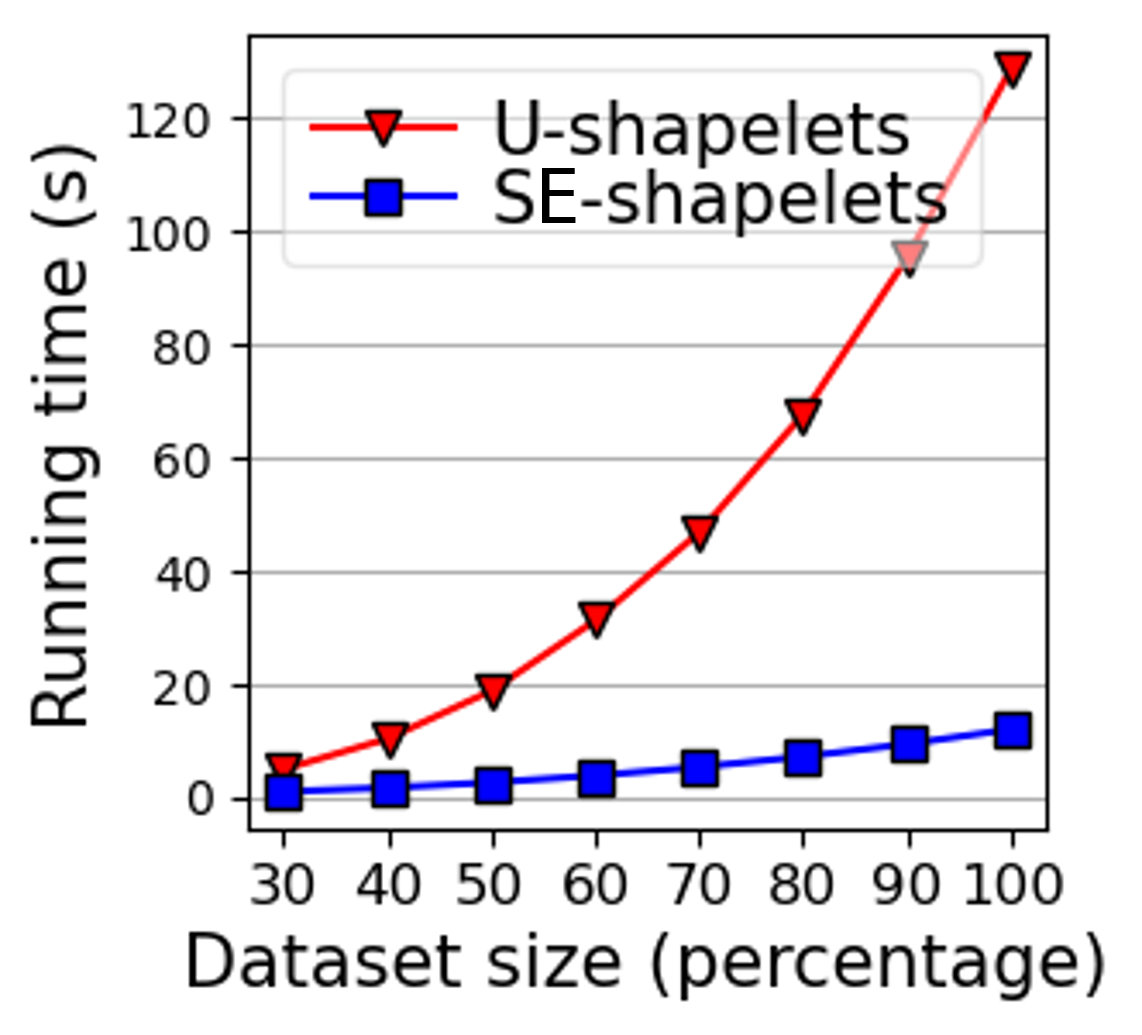}}\hspace{0.07in}
\subfloat[Coffee]{\includegraphics[width=1.63in]{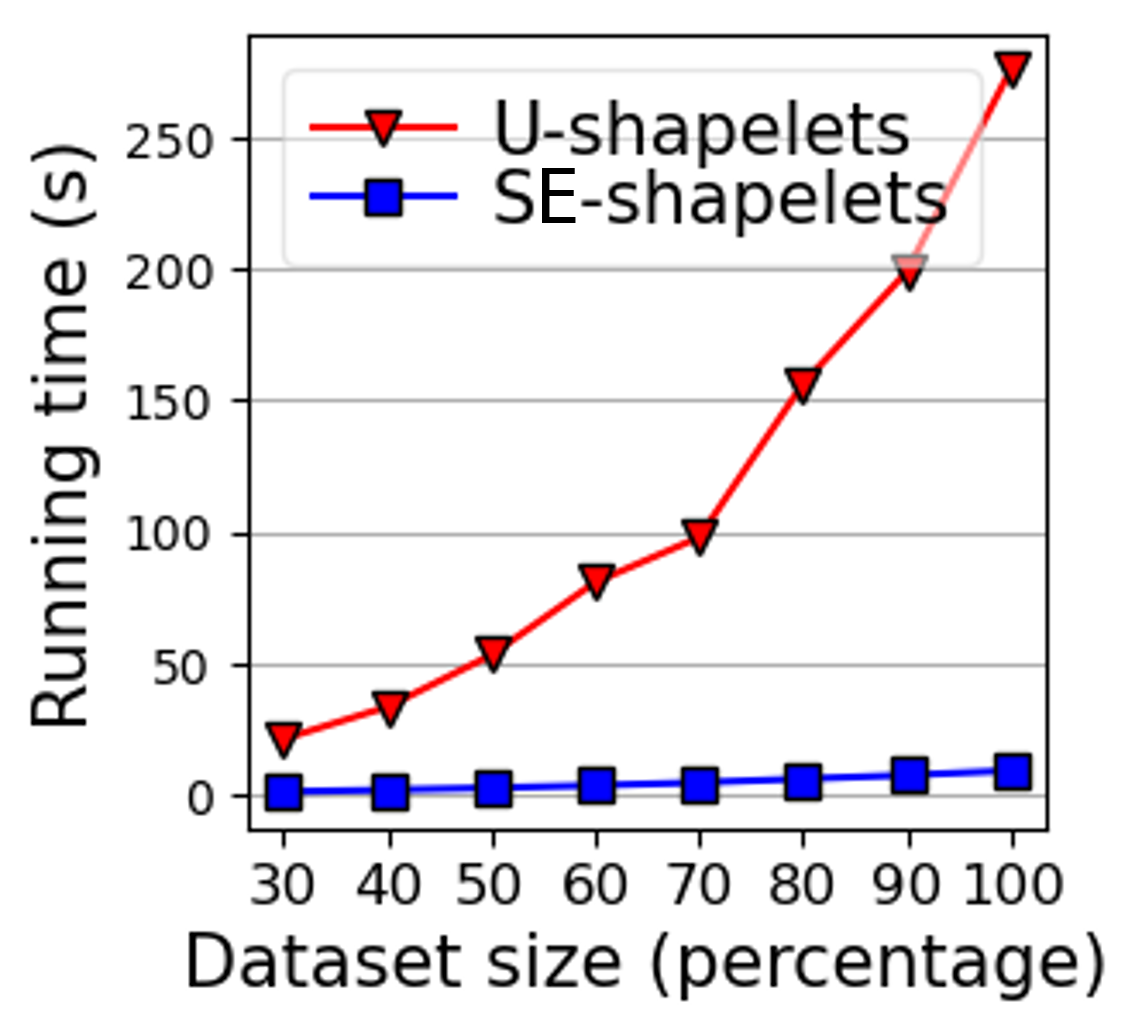}}\hspace{0.07in}
\subfloat[CBF]{\includegraphics[width=1.60in]{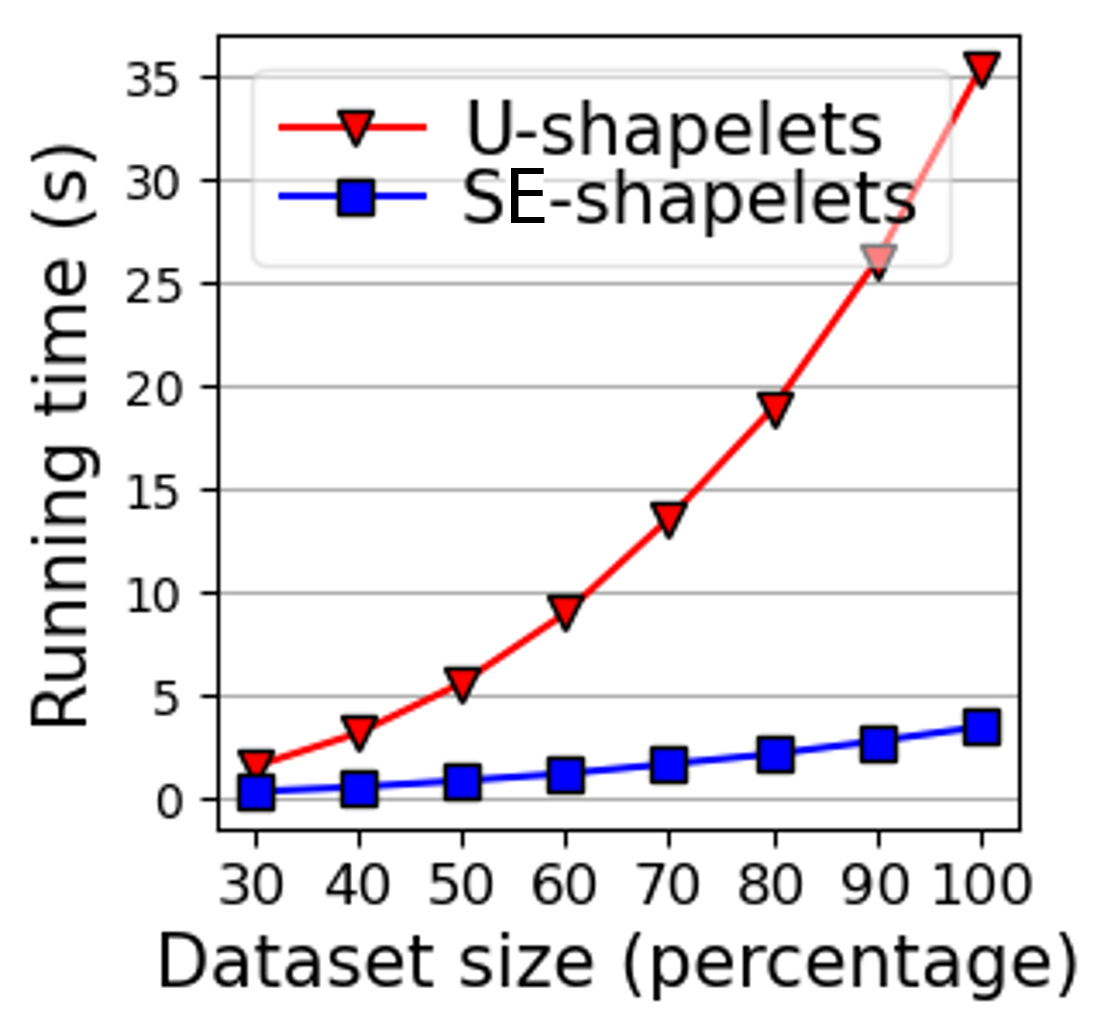}}\hspace{0.07in}
\subfloat[TwoLeadECG]{\includegraphics[width=1.60in]{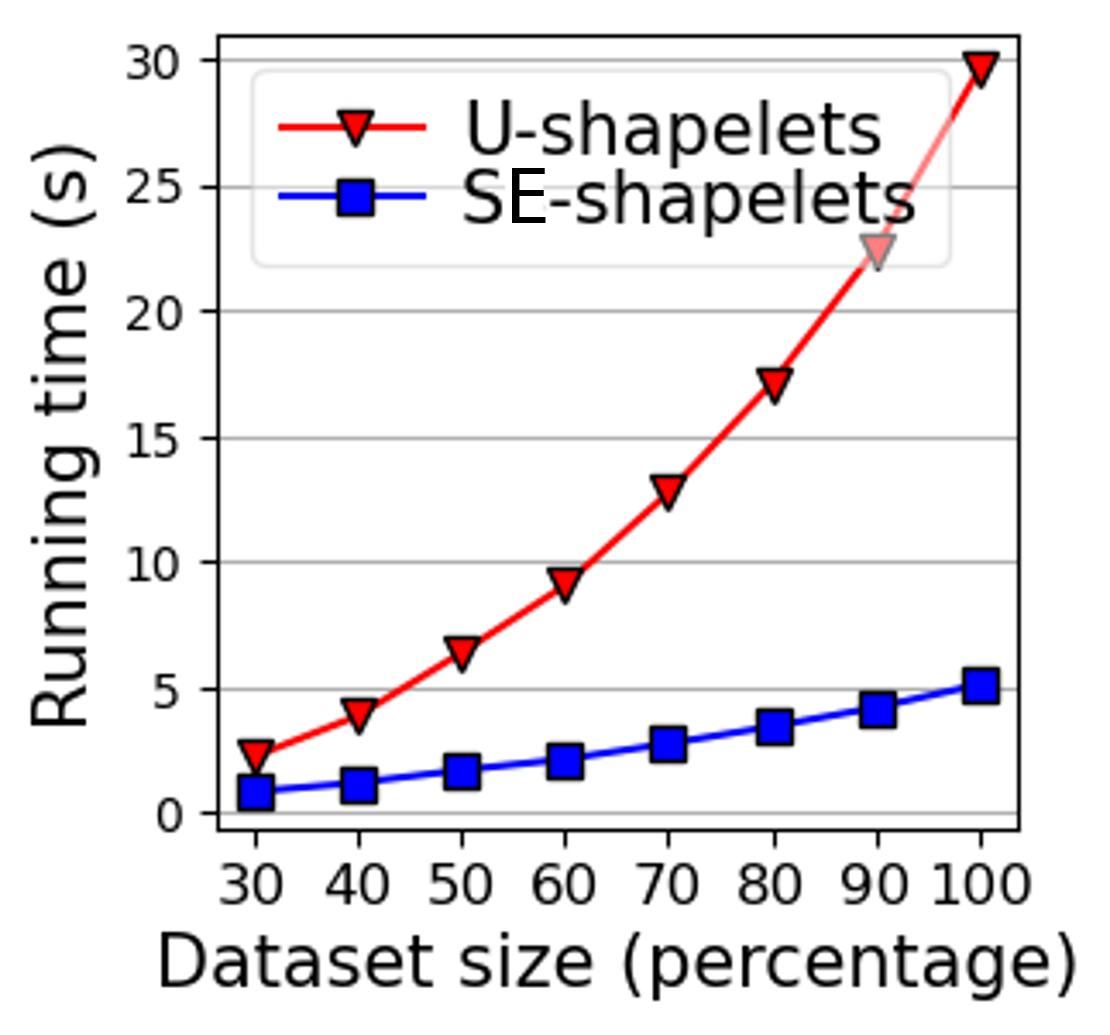}} \\
\caption{Running time of SE-shapelets and U-shapelets with respect to different dataset sizes.}
\label{fig:runt}
\end{figure*}

We show the distribution of the RI improvement on the 85 datasets in Fig. \ref{fig:distribution}.
The results show that, with slight supervision, in many datasets the clustering accuracy is increased by more than $10\%$, and the largest increase is over 100\% (from 0.497 to 1.000 in Coffee dataset). For the few datasets that SE-shapelets obtain accuracy lower than U-shapelets, the decreases of accuracy are all small and the largest decrease is around 4\% (from 0.568 to 0.545 in ScreenType dataset).

We further use the results from four case datasets (BeetleFly, Coffee, CBF and TwoLeadECG) to show the shapelets discovered by SE-shapelets are more effective than those discovered by U-shapelets. 
With the optimal shapelets, the four datasets are mapped to the distance-to-shapelets representation, and their distances are visualized in 2D space using TSNE \citep{tsne} as shown in Fig. \ref{fig:tsne}, respectively.

In BeetleFly (Fig. \ref{fig:tsne} (a)), Coffee (Fig. \ref{fig:tsne} (b)) and TwoLeadECG (Fig. \ref{fig:tsne} (d)) datasets, the shapelets discovered by U-shapelets  split time series into two groups but each group contains time series of different classes (the bottom row in Fig. \ref{fig:tsne}).
However, as shown in the top row in Fig. \ref{fig:tsne}, with slight supervision (e.g., only 2 labeled time series for Coffee), SE-shapelets discovers shapelets that effectively split time series into two groups that mostly contain time series of the same class.
For CBF dataset, although U-shapelets roughly distinguish time series of different classes, SE-shapelets can further improve the performance, especially for the two classes represented by the red circle and the blue square, respectively.

\subsection{SE-shapelets vs Other Unsupervised Methods}
We also compare SE-shapelets with three strong unsupervised time series clustering methods to show the effectiveness. These unsupervised methods are AutoShape, Kshape, and TS3C. Specifically, \textbf{AutoShape} \citep{autoshape} adopts objective optimization to learn shapelets as latent representation learning for time series clustering, under the framework of autoencoder. \textbf{Kshape} \citep{kshape} approaches by a marriage of shape-based distance with spectral analysis to address time series distortions. \textbf{TS3C} \citep{ts3c} discovers multiple subsequence clusters to represent the topological structure of time series, based on which similar time series are grouped. 

\begin{figure}[!htbp]
\centering
\subfloat{\includegraphics[width=3.3in]{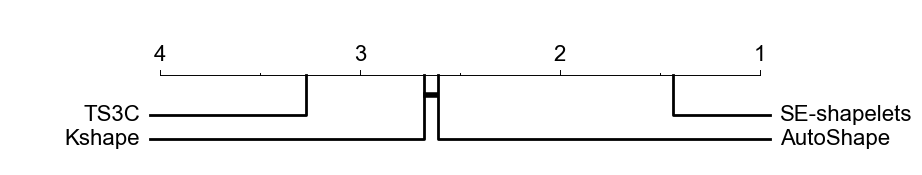}}
\caption{Ranks (lower is better) of the SE-shapelets and unsuperivsed methods on the UCR datasets.}
\label{fig:cdd}
\end{figure}

As the summarized results in Table \ref{tab:unsuper} show, SE-shapelets achieves the best clustering accuracy (in average) on the UCR time series datasets, which is 10.3\% higher than the second best result (achieved by AutoShape, 0.705). Meanwhile, the shapelet-based AutoShape outperforms both Kshape and TS3C, and that again demonstrates the effectiveness of using shapelet for time series clustering. In addition, SE-shapelets also obtains the best Wilcoxon test ranking, as shown in Fig. \ref{fig:cdd}, and that means its performance is statistical better than compared unsupervised methods. 

\begin{table}[h]
\centering
  \caption{Accuracy comparison between SE-shapelets and unsupervised time series clustering methods (with the best in bold).}
  \begin{tabular}{lcccc}
  \hline 
  &SE-shapelets &AutoShape &Kshape &TS3C \\
  \hline
  Avg. RI	  &\textbf{0.778}	&0.705 &0.690 &0.66 \\ 
  Avg. Rank &\textbf{1.4}	    &2.6   &2.7   &3.3  \\ 
    \hline
  \end{tabular}
\label{tab:unsuper}
\end{table}

\subsection{Parameter Study}
We further study the influence of input parameters of SE-shapelets on its performance. SE-shapelets requires three parameters, i.e., the number of shapelets ($k$), the length of shapelets ($\hat{l}$), and the weight of variance ($\lambda$). Since the clustering accuracy is not sensitive to $\lambda$, we fix it as $0.1$ in this experiment. We vary $k$ from $2$ to $9$ and $\hat{l}$ from $0.01l$ to $0.11l$, where $l$ is the length of time series in the datasets. The results of the clustering accuracy under different parameters are shown in Fig. \ref{fig:param}.

In general, the clustering accuracy does not necessarily increase with more shapelets; for example, in BeetleFly (Fig. \ref{fig:param} (a)) and CBF (Fig. \ref{fig:param} (c)), the best accuracy is achieved with 6 shapelets and 2 shapelets, respectively. This observation again shows that time series of these datasets can be properly discriminated with only a small number of subsequences. Meanwhile, the clustering accuracy roughly increases with the length of shapelets in the four datasets (Fig. \ref{fig:param} (a-d)), mainly because too short shapelets cannot fully capture meaningful local temporal patterns; while the best clustering accuracy still is achieved with relatively short shapelets, i.e., $\hat{l}\leq 0.1l$.

\subsection{Running Time}
We show the running time efficiency of SE-shapelets by comparing it with U-shapelets, regarding varying dataset sizes on four datasets. SE-shapelets and U-shapelets are run 10 times to obtain the average running time, with $k=3$, $\hat{l}=0.1l$. The results are shown in Fig. \ref{fig:runt}.

The results show that SE-shapelets is more scalable than U-shapelets on all four datasets. Specifically, the running time of SE-shapelets increases much slower than that of U-shapelets with larger dataset sizes, because SE-shapelets can avoid a large number of uninformative shapelet candidates.
Although the complexity of $FindChain$ is quadratic to time series length, $FindChain$ is only called for a small amount of labeled/pseudo-labeled time series and therefore does not increase running time in a quadratic manner. In real-life applications, it is rather convenient to implement multi-thread processes to call $FindChain$ in parallel for further acceleration. For the Spectral clustering used in SE-shapelets, there are many scalable versions \citep{spectral1,spectral2} and we use an efficient implementation from sklearn\footnote{https://scikit-learn.org/stable/}.

\section{Conclusion}
This paper proposes an SE-shapelets method that discovers representative shapelets with a small number of labeled and propagated pseudo-labeled time series for accurate time series clustering. SE-shapelets defines a new $SSC$ to extract salient subsequences (as candidate shapelets) from a label/pseudo-labeled time series, and develops an effective $LDS$ algorithm to discover representative shapelets suitable for time series clustering. Through extensive evaluation on UCR time series datasets, we show that SE-shapelets generally achieves better clustering accuracy than counterpart semi-supervised time series clustering methods. Furthermore, we demonstrate that, with a small number of labels, shapelets discovered by SE-shapelets are better at clustering time series than shapelets discovered by the unsupervised method.

\section*{Acknowledgments}

This work was supported in part by the Australian Research Council under Grant LP190100594 and DE140100387.

\bibliographystyle{apalike}

\bibliography{ref}

\begin{thebibliography}{}

\bibitem[Abbasimehr and Bahrini, 2022]{process}
Abbasimehr, H. and Bahrini, A. (2022).
\newblock An analytical framework based on the recency, frequency, and monetary
  model and time series clustering techniques for dynamic segmentation.
\newblock {\em Expert Systems with Applications}, 192:116373.

\bibitem[Alcock et~al., 1999]{synthetic}
Alcock, R.~J., Manolopoulos, Y., et~al. (1999).
\newblock Time-series similarity queries employing a feature-based approach.
\newblock In {\em 7th Hellenic Conference on Informatics}, pages 27--29.

\bibitem[Amouri et~al., 2023]{cdps}
Amouri, H.~E., Lampert, T., Gan{\c{c}}arski, P., and Mallet, C. (2023).
\newblock Cdps: Constrained dtw-preserving shapelets.
\newblock In {\em Machine Learning and Knowledge Discovery in Databases:
  European Conference, ECML PKDD 2022, Grenoble, France, September 19--23,
  2022, Proceedings, Part I}, pages 21--37. Springer.

\bibitem[Balakrishnama and Ganapathiraju, 1998]{lda}
Balakrishnama, S. and Ganapathiraju, A. (1998).
\newblock Linear discriminant analysis-a brief tutorial.
\newblock {\em Institute for Signal and Information Processing}, 18(1998):1--8.

\bibitem[Basu et~al., 2002]{seedkmeans}
Basu, S., Banerjee, A., and Mooney, R. (2002).
\newblock Semi-supervised clustering by seeding.
\newblock In {\em In Proceedings of 19th International Conference on Machine
  Learning (ICML)}. Citeseer.

\bibitem[Cai et~al., 2021]{minidtw}
Cai, B., Huang, G., Samadiani, N., Li, G., and Chi, C.-H. (2021).
\newblock Efficient time series clustering by minimizing dynamic time warping
  utilization.
\newblock {\em IEEE Access}, 9:46589--46599.

\bibitem[Cai et~al., 2020]{stopword}
Cai, B., Huang, G., Xiang, Y., Angelova, M., Guo, L., and Chi, C.-H. (2020).
\newblock Multi-scale shapelets discovery for time-series classification.
\newblock {\em International Journal of Information Technology \& Decision
  Making}, 19(03):721--739.

\bibitem[Chen et~al., 2021]{wear}
Chen, L., Liu, X., Peng, L., and Wu, M. (2021).
\newblock Deep learning based multimodal complex human activity recognition
  using wearable devices.
\newblock {\em Applied Intelligence}, 51(6):4029--4042.

\bibitem[Chen and Ng, 2004]{edit}
Chen, L. and Ng, R. (2004).
\newblock On the marriage of lp-norms and edit distance.
\newblock In {\em Proceedings of the Thirtieth international conference on Very
  large data bases-Volume 30}, pages 792--803.

\bibitem[Chen et~al., 2015]{ucrts}
Chen, Y., Keogh, E., Hu, B., Begum, N., Bagnall, A., Mueen, A., and Batista, G.
  (2015).
\newblock The ucr time series classification archive.
\newblock \url{www.cs.ucr.edu/~eamonn/time_series_data/}.

\bibitem[Dau et~al., 2016]{semidtw}
Dau, H.~A., Begum, N., and Keogh, E. (2016).
\newblock Semi-supervision dramatically improves time series clustering under
  dynamic time warping.
\newblock In {\em Proceedings of the 25th ACM International on Conference on
  Information and Knowledge Management}, pages 999--1008.

\bibitem[Davidson and Ravi, 2005]{COP-Kmeans}
Davidson, I. and Ravi, S. (2005).
\newblock Clustering with constraints: Feasibility issues and the k-means
  algorithm.
\newblock In {\em Proceedings of the 2005 SIAM international conference on data
  mining}, pages 138--149. SIAM.

\bibitem[Ding et~al., 2008]{1nn}
Ding, H., Trajcevski, G., Scheuermann, P., Wang, X., and Keogh, E. (2008).
\newblock Querying and mining of time series data: experimental comparison of
  representations and distance measures.
\newblock {\em Proceedings of the VLDB Endowment}, 1(2):1542--1552.

\bibitem[Ding et~al., 2019]{longest}
Ding, J., Fang, J., Zhang, Z., Zhao, P., Xu, J., and Zhao, L. (2019).
\newblock Real-time trajectory similarity processing using longest common
  subsequence.
\newblock In {\em 2019 IEEE 21st International Conference on High Performance
  Computing and Communications; IEEE 17th International Conference on Smart
  City; IEEE 5th International Conference on Data Science and Systems
  (HPCC/SmartCity/DSS)}, pages 1398--1405. IEEE.

\bibitem[Ding et~al., 2015]{yading}
Ding, R., Wang, Q., Dang, Y., Fu, Q., Zhang, H., and Zhang, D. (2015).
\newblock Yading: fast clustering of large-scale time series data.
\newblock {\em Proceedings of the VLDB Endowment (VLDB 2015)}, 8(5):473--484.

\bibitem[Fotso et~al., 2020]{fots}
Fotso, V. S.~S., Nguifo, E.~M., and Vaslin, P. (2020).
\newblock Frobenius correlation based u-shapelets discovery for time series
  clustering.
\newblock {\em Pattern Recognition}, page 107301.

\bibitem[Grabocka et~al., 2014]{lts}
Grabocka, J., Schilling, N., Wistuba, M., and Schmidt-Thieme, L. (2014).
\newblock Learning time-series shapelets.
\newblock In {\em Proceedings of the 20th ACM SIGKDD International Conference
  on Knowledge Discovery and Data Mining (KDD 2014)}, pages 392--401. ACM.

\bibitem[Grabocka et~al., 2016]{sd}
Grabocka, J., Wistuba, M., and Schmidt-Thieme, L. (2016).
\newblock Fast classification of univariate and multivariate time series
  through shapelet discovery.
\newblock {\em Knowledge and Information Systems}, 49(2):429--454.

\bibitem[Guijo-Rubio et~al., 2020]{ts3c}
Guijo-Rubio, D., Dur{\'a}n-Rosal, A.~M., Guti{\'e}rrez, P.~A., Troncoso, A.,
  and Herv{\'a}s-Mart{\'\i}nez, C. (2020).
\newblock Time-series clustering based on the characterization of segment
  typologies.
\newblock {\em IEEE Transactions on Cybernetics}.

\bibitem[He et~al., 2021]{fsskmeans}
He, G., Pan, Y., Xia, X., He, J., Peng, R., and Xiong, N.~N. (2021).
\newblock A fast semi-supervised clustering framework for large-scale time
  series data.
\newblock {\em IEEE Transactions on Systems, Man, and Cybernetics: Systems},
  51(7):4201--4216.

\bibitem[He and Tan, 2018]{sensory1}
He, H. and Tan, Y. (2018).
\newblock Pattern clustering of hysteresis time series with multivalued mapping
  using tensor decomposition.
\newblock {\em IEEE Transactions on Systems, Man, and Cybernetics: Systems},
  48(6):993--1004.

\bibitem[He et~al., 2022]{uav1}
He, K., Yu, D., Wang, D., Chai, M., Lei, S., and Zhou, C. (2022).
\newblock Graph attention network-based fault detection for uavs with
  multivariant time series flight data.
\newblock {\em IEEE Transactions on Instrumentation and Measurement}, 71:1--13.

\bibitem[Holder et~al., 2023]{elastic}
Holder, C., Middlehurst, M., and Bagnall, A. (2023).
\newblock A review and evaluation of elastic distance functions for time series
  clustering.
\newblock {\em Knowledge and Information Systems}, pages 1--45.

\bibitem[Holm, 1979]{cd}
Holm, S. (1979).
\newblock A simple sequentially rejective multiple test procedure.
\newblock {\em Scandinavian Journal of Statistics}, pages 65--70.

\bibitem[Hu et~al., 2022]{intro_radar}
Hu, Y., Jia, X., Tomizuka, M., and Zhan, W. (2022).
\newblock Causal-based time series domain generalization for vehicle intention
  prediction.
\newblock In {\em 2022 International Conference on Robotics and Automation
  (ICRA)}, pages 7806--7813. IEEE.

\bibitem[Huang et~al., 2019]{spectral1}
Huang, D., Wang, C.-D., Wu, J.-S., Lai, J.-H., and Kwoh, C.-K. (2019).
\newblock Ultra-scalable spectral clustering and ensemble clustering.
\newblock {\em IEEE Transactions on Knowledge and Data Engineering},
  32(6):1212--1226.

\bibitem[Jiang et~al., 2022]{cssc}
Jiang, Z., Zhan, Y., Mao, Q., and Du, Y. (2022).
\newblock Semi-supervised clustering under a compact-cluster assumption.
\newblock {\em IEEE Transactions on Knowledge and Data Engineering}.

\bibitem[Karlsson et~al., 2016]{random}
Karlsson, I., Papapetrou, P., and Bostr{\"o}m, H. (2016).
\newblock Generalized random shapelet forests.
\newblock {\em Data Mining and Knowledge Discovery}, 30(5):1053--1085.

\bibitem[Lelis and Sander, 2009]{ssdbscan}
Lelis, L. and Sander, J. (2009).
\newblock Semi-supervised density-based clustering.
\newblock In {\em 2009 Ninth IEEE International Conference on Data Mining},
  pages 842--847. IEEE.

\bibitem[Li et~al., 2022]{autoshape}
Li, G., Choi, B., Xu, J., Bhowmick, S.~S., Mah, D. N.-y., and Wong, G. L.-H.
  (2022).
\newblock Autoshape: An autoencoder-shapelet approach for time series
  clustering.
\newblock {\em arXiv preprint arXiv:2208.04313}.

\bibitem[Li et~al., 2023]{survey}
Li, H., Liu, Z., and Wan, X. (2023).
\newblock Time series clustering based on complex network with synchronous
  matching states.
\newblock {\em Expert Systems with Applications}, 211:118543.

\bibitem[Maaten and Hinton, 2008]{tsne}
Maaten, L. v.~d. and Hinton, G. (2008).
\newblock Visualizing data using t-sne.
\newblock {\em Journal of Machine Learning Research}, 9:2579--2605.

\bibitem[Paparrizos and Gravano, 2017]{kshape}
Paparrizos, J. and Gravano, L. (2017).
\newblock Fast and accurate time-series clustering.
\newblock {\em ACM Transactions on Database Systems}, 42:1--49.

\bibitem[Petitjean et~al., 2011]{kdba}
Petitjean, F., Ketterlin, A., and Gan{\c{c}}arski, P. (2011).
\newblock A global averaging method for dynamic time warping, with applications
  to clustering.
\newblock {\em Pattern Recognition}, 44(3):678--693.

\bibitem[Schneider and Shiffrin, 1977]{salience}
Schneider, W. and Shiffrin, R.~M. (1977).
\newblock Controlled and automatic human information processing: I. detection,
  search, and attention.
\newblock {\em Psychological Review}, 84(1):1.

\bibitem[Shi and Malik, 2000]{spectral}
Shi, J. and Malik, J. (2000).
\newblock Normalized cuts and image segmentation.
\newblock {\em IEEE Transactions on Pattern Analysis and Machine Intelligence},
  22(8):888--905.

\bibitem[Van~Craenendonck et~al., 2018]{COBRAS}
Van~Craenendonck, T., Meert, W., Duman{\v{c}}i{\'c}, S., and Blockeel, H.
  (2018).
\newblock Cobras ts: A new approach to semi-supervised clustering of time
  series.
\newblock In {\em International Conference on Discovery Science}, pages
  179--193. Springer.

\bibitem[Van~Lierde et~al., 2019]{spectral2}
Van~Lierde, H., Chow, T.~W., and Chen, G. (2019).
\newblock Scalable spectral clustering for overlapping community detection in
  large-scale networks.
\newblock {\em IEEE Transactions on Knowledge and Data Engineering},
  32(4):754--767.

\bibitem[Wang et~al., 2019]{unlabeled}
Wang, H., Zhang, Q., Wu, J., Pan, S., and Chen, Y. (2019).
\newblock Time series feature learning with labeled and unlabeled data.
\newblock {\em Pattern Recognition}, 89:55--66.

\bibitem[Wei and Keogh, 2006]{pseudo}
Wei, L. and Keogh, E. (2006).
\newblock Semi-supervised time series classification.
\newblock In {\em Proceedings of the 12th ACM SIGKDD International Conference
  on Knowledge Discovery and Data Mining}, pages 748--753.

\bibitem[Yamaguchi et~al., 2022]{learning1}
Yamaguchi, A., Ueno, K., and Kashima, H. (2022).
\newblock Learning time-series shapelets enhancing discriminability.
\newblock In {\em Proceedings of the 2022 SIAM International Conference on Data
  Mining (SDM)}, pages 190--198. SIAM.

\bibitem[Ye and Keogh, 2009]{gain0}
Ye, L. and Keogh, E. (2009).
\newblock Time series shapelets: a new primitive for data mining.
\newblock In {\em Proceedings of the 15th ACM SIGKDD International Conference
  on Knowledge Discovery and Data Mining}, pages 947--956.

\bibitem[Yin et~al., 2022]{sensor}
Yin, C., Zhang, S., Wang, J., and Xiong, N.~N. (2022).
\newblock Anomaly detection based on convolutional recurrent autoencoder for
  iot time series.
\newblock {\em IEEE Transactions on Systems, Man, and Cybernetics: Systems},
  52(1):112--122.

\bibitem[Yu et~al., 2022]{seg}
Yu, Q., Wang, H., Kim, D., Qiao, S., Collins, M., Zhu, Y., Adam, H., Yuille,
  A., and Chen, L.-C. (2022).
\newblock Cmt-deeplab: Clustering mask transformers for panoptic segmentation.
\newblock In {\em Proceedings of the IEEE/CVF Conference on Computer Vision and
  Pattern Recognition}, pages 2560--2570.

\bibitem[Yu et~al., 2017]{compact}
Yu, S., Yan, Q., and Yan, X. (2017).
\newblock Improving u-shapelets clustering performance: An shapelets quality
  optimizing method.
\newblock {\em International Journal of Hybrid Information Technology},
  10(4):27--40.

\bibitem[Zakaria et~al., 2012]{ushape}
Zakaria, J., Mueen, A., and Keogh, E. (2012).
\newblock Clustering time series using unsupervised-shapelets.
\newblock In {\em IEEE 12th International Conference on Data Mining (ICDM
  2012)}, pages 785--794. IEEE.

\bibitem[Zhang and Sun, 2022]{learning2}
Zhang, N. and Sun, S. (2022).
\newblock Multiview unsupervised shapelet learning for multivariate time series
  clustering.
\newblock {\em IEEE Transactions on Pattern Analysis and Machine Intelligence},
  45(4):4981--4996.

\bibitem[Zhang et~al., 2018]{ussl}
Zhang, Q., Wu, J., Zhang, P., Long, G., and Zhang, C. (2018).
\newblock Salient subsequence learning for time series clustering.
\newblock {\em IEEE Transactions on Pattern Analysis and Machine Intelligence},
  41(9):2193--2207.

\bibitem[Zhou et~al., 2015]{ssncut}
Zhou, J., Zhu, S.-F., Huang, X., and Zhang, Y. (2015).
\newblock Enhancing time series clustering by incorporating multiple distance
  measures with semi-supervised learning.
\newblock {\em Journal of Computer Science and Technology}, 30(4):859--873.

\end{thebibliography}

\end{document}